\documentclass[11pt]{article}
\usepackage{fullpage}
\usepackage{wustyle}
\usepackage{natbib}
\usepackage{indentfirst}
\usepackage{blkarray}
\usepackage{authblk}
\usepackage{braket}
\usepackage{outlines}
\usepackage{float}
\usepackage[a4paper,margin=1in]{geometry}
\usepackage{todonotes}

\newcommand{\uo}{^{(1)}}
\newcommand{\ut}{^{(2)}}
\newcommand{\uk}{^{(k)}}
\newcommand{\sumM}{\sum_{j=1}^M}

\def\trans{^{\scriptscriptstyle \sf T}}

\definecolor{red2}{rgb}{0.7, 0, 0.1}
\usepackage{color}

\title{Analyzing the Role of Permutation Invariance in \\ Linear Mode Connectivity}
\author[1]{Keyao Zhan$^*$}
\author[2]{Puheng Li\thanks{The first two authors contributed equally.}}
\author[3,4]{Lei Wu\thanks{Corresponding author (\texttt{leiwu@math.pku.edu.cn}).}}
\affil[1]{Department of Biostatistics, Harvard T.H. Chan School of Public Health.}
\affil[2]{Department of Statistics, Stanford University}
\affil[3]{School of Mathematical Sciences, Peking University}
\affil[4]{Center for Machine Learning Research, Peking University}
\date{\vspace*{-2em}}

\begin{document}

\maketitle

\thispagestyle{empty}

\begin{abstract}

It was empirically observed in \cite{entezari2021role}  that when accounting for the permutation invariance of neural networks, there is likely no loss barrier along the linear interpolation between two SGD solutions -- a phenomenon known as linear mode connectivity (LMC) modulo permutation.
  This phenomenon has sparked significant attention due to both its theoretical interest and practical relevance in applications such as model merging. In this paper, we provide a fine-grained analysis of this phenomenon for two-layer ReLU networks under a teacher-student setup. We show that as the student network width $m$ increases,  the LMC loss barrier modulo permutation exhibits a {\bf double descent} behavior. Particularly, when $m$ is sufficiently large, the barrier decreases to zero at a rate $O(m^{-1/2})$. Notably,  this rate does not suffer from the curse of dimensionality and demonstrates how substantial permutation can reduce the LMC loss barrier. Moreover, we observe a sharp transition in the sparsity of GD/SGD solutions when increasing the learning rate and investigate how this sparsity preference affects the LMC loss barrier modulo permutation. Experiments on both synthetic and MNIST datasets corroborate our theoretical predictions and reveal a similar trend for more complex network architectures.

\end{abstract}




\section{Introduction}
Despite the remarkable successes of modern deep neural networks, a theoretical understanding of their underlying mechanisms remains elusive. One major challenge in deep learning theory is uncovering the structure of the high-dimensional loss landscape \citep{wu2017towards,li2018visualizing,zhang2021embedding,mei2018mean}, which is crucial for understanding the training dynamics. Unfortunately, the loss landscape exhibits significant complexity, high dimensionality, and degeneracy, characterized by numerous minima, symmetries, and saddle points \citep{zhang2021understanding, draxler2018essentially}. For instance,  over-parameterized networks have the capacity to represent multiple functions that achieve similar performance on training data but differ significantly in parameter space \citep{neyshabur2017exploring,li2018benefit,liu2022loss}. Additionally, inherent scale and permutation invariances of neural networks allow a single function to be expressed through different parameter configurations within the same network. Despite these challenges, it is believed that the loss landscape encountered during practical training  possess intricate yet benign property that facilitate the effectiveness of gradient-based optimization \citep{ge2016matrix, keskar2017on}.

A particularly intriguing phenomenon uncovered in recent work is Mode Connectivity \citep{freeman2017topology, draxler2018essentially, garipov2018loss}: Different optima found by independent gradient-based optimization runs turn out to be connected, i.e., there exists a path connecting them along which the loss or accuracy remains nearly constant. This is surprising because one would expect distinct optima of a non-convex function to lie in separate, isolated valleys--yet this separation does not occur in practice.

More recently, a stronger form of mode connectivity known as Linear Mode Connectivity (LMC) was proposed \citep{frankle2020linear}: different optima can be connected by a linear path that does not pass through any loss barrier. LMC has since been used to explore various deep learning phenomena, such as generalization \citep{juneja2022linear}. While LMC usually does not emerge between two independently trained networks, it consistently appears in the following sense \citep{entezari2021role, ainsworth2022git}: For two independently trained global minima, there is a neuron permutation of one global minimum that makes it linearly connected to the other once the permutation is applied. However, despite these observations, this remains a conjecture awaiting a solid theoretical explanation.

\paragraph*{Our contribution.}
In this paper,
we provide a  theoretical explanation of LMC modulo permutation for two-layer ReLU networks under a teacher-student setup~\citep{lin2024exploring}. This setup allows us to quantify the influence of permutation invariance for LMC. Let $m$ and $M$ denote the numbers of neurons of the student and teacher networks, respectively. Our contributions are summarized as follows:
\begin{itemize}
   \item 
   First, we prove that applying permutations can {\bf substantially} reduce the loss barrier of LMC, compared to direct linear interpolation without permutation. Specifically, the loss barrier modulo permutation diminishes to zero  at a  rate $O(m^{-1/2})$ when  $m$ is sufficiently large--crucially independent of the input dimension, thereby avoiding the curse of dimensionality. This offers a quantitative perspective on how permutations enhance LMC by lowering the loss barrier. In contrast, the upper bound  $O(m^{-1/(2d+4)})$ in~\cite{entezari2021role} (where $d$ is the input dimension) suffers from the curse of dimensionality.

   \item Second, we provide a theoretical explanation for the ``peak phenomenon'' observed in \cite{entezari2021role}, where the loss barrier (modulo permutation) initially increases and then decreases to zero as network width increases. We pinpoint the exact location of this peak in our setup. Moreover, we identify a {\bf double descent} \citep{belkin2019reconciling} behavior in the LMC loss barrier (modulo permutation): it decreases as $m$ increases up to $m=M$, then rises to a peak at $m=2M$, before finally decreasing again.

   \item  Third, we observe a sharp transition in the sparsity of GD/SGD solutions as the learning rate increases, and we further investigate how this preference for sparse solutions impacts the LMC modulo permutation. 
\end{itemize}


\subsection{Related Works}

\paragraph*{Mode connectivity.}
In the initial work  \cite{freeman2017topology}, mode connectivity was proved for both linear networks and two-layer ReLU networks with $\ell_2$ regularization. \cite{garipov2018loss} and \cite{fort2019large} empirically discovered the piecewise-linear connecting paths. \cite{nagarajan2019uniform} first observed Linear Mode Connectivity (LMC), i.e., the near-constant-loss connecting path can be linear, on models trained on MNIST starting from the same random initialization. Later, \cite{frankle2020linear} observed LMC in more difficult datasets, for networks that are jointly trained for a short period of time before going through independent training. \cite{fort2020deep} explore the connection between LMC and the Neural Tangent Kernel dynamics. \citep{liang2018understanding, kuditipudi2019explaining, nguyen2019connected, nguyen2018loss} provide some great insights into the geometry and connectivity of the loss landscape. More recent advancements, such as \citep{zhou2023going, ferbach2024proving}, introduce new perspectives on LMC, including layerwise connectivity and optimal transport approaches. 
 
\paragraph*{Permutation invariance for LMC.} As a conjecture proposed by \cite{entezari2021role}, permutation invariance of linear mode connectivity has been constantly studied in recent works. \cite{benzing2022random} provided a simple algorithm for finding the optimal permutation and showed its connection with generalization. \cite{ainsworth2022git} showed that the global minima fall into a connected low-loss basin after permutation. \cite{jordan2022repair} explored the limits of permutation that, in some regimes, permutation brings little improvement to linear mode connectivity.

\section{Preliminaries}

\noindent \textbf{Notation.} For $n\in \mathbb{N}$, let $[n]=\{1,2, \ldots, n\}$. For a compact set $\Omega$, denote by $\text{Unif}(\Omega)$ the uniform distribution over $\Omega$. Let $\left\{e_j\right\}_{j=1}^d$ be the canonical basis of $\mathbb{R}^d$. Let $\mathbb{S}^{d-1}=\left\{x \in \mathbb{R}^d:\|x\|_2=1\right\}$ and $\tau_{d-1}=\operatorname{Unif}\left(\mathbb{S}^{d-1}\right)$.

Denote by $\mathcal{X}$ the input space, $\mathcal{Y}$ the output space, and $\mathcal{D}$ denote a data distribution over $\mathcal{X} \times \mathcal{Y}$. 
Let $f: \mathcal{X} \times \Theta \mapsto \mathcal{Y}$ be a neural network with $\Theta$ denoting the parameter space. For a given loss function $\ell: \mathcal{Y} \times \mathcal{Y} \mapsto \mathbb{R}$, the corresponding loss landscape is determined by 
\begin{equation}
    \mathcal{L}(\theta)=\mathbb{E}_{(x, y) \sim \mathcal{D}}[\ell(f(x ; \theta), y)].
\end{equation}
In this paper, we focus on the over-realization regime where $\inf _{\theta \in \Theta} \mathcal{L}(\theta)=0$. Then, the global minima manifold is given by
\begin{equation}
    \mathcal{M}=\{\theta \in \Theta: \mathcal{L}(\theta)=0\} .
\end{equation}
Given that the parameter on the linear interpolation of two global minima $\theta_1$ and $\theta_2$ is not necessarily a global minima, we define the \textbf{linear barrier} between two minima \citep{frankle2020linear} as
\begin{align*}
    B\left(\theta_1, \theta_2\right)=\sup_{\lambda \in [0,1]} \left[\mathcal{L}\left(\lambda \theta_1+(1-\lambda) \theta_2\right)\right]-\left[\lambda \mathcal{L}\left(\theta_1\right)+(1-\lambda) \mathcal{L}\left(\theta_2\right)\right].
\end{align*}
When empirically evaluated, we approximate this by evaluating $\lambda \in[0,1]$ with a step size 0.1 , calculating the barrier for $\lambda=0.0,0.1,0.2, \cdots, 1.0$ and taking the maximum value. We find in practice, this discretization is always sufficient, as the landscape along the linear interpolation is generally smooth and well-behaved. See Figure \ref{interpolation} for a few illustrations. 

\noindent \textbf{Model setup.} We consider the two-layer ReLU network under the teacher-student setting, where the label is generated by a teacher network: $f^*(x)=\sum_{j=1}^M \sigma(w_j^* \cdot x)$ and the activation function is ReLU function defined by $\sigma(z)=$ $\max (0, z)$. The number of teacher neurons is $M$, and the dimension of input is $d$, that is, $x \in \mathbb{R}^d$. We make the following assumption under this network architecture:

\begin{assumption} 
Suppose $M \leqslant d, w_j^*=e_j$ for $j \in [M]$, and $x \sim \tau_{d-1}$.
\label{assum:orthonormal}
\end{assumption} 
By the rotational symmetry, this specific assumption is equivalent to only assuming $\{w_j^*\}_{j=1}^{M}$ to be orthonormal. In such a case, the loss objective function is 
\begin{equation}
    L(\theta)=\mathbb{E}_{x \sim \tau_{d-1}}\left[\left(\sum_{i=1}^m \sigma\left(w_i \cdot x\right)-\sum_{j=1}^M \sigma\left(x_j\right)\right)^2\right],
\label{eq:lossfun}
\end{equation}
where $m$ denotes the number of neurons of the student network and $\theta=\left(w_1, w_2, \ldots, w_m\right)\trans=\left(w_{i, j}\right) \in$ $\mathbb{R}^{m \times d}$. Using this notation, each row of $W$ represents a student neuron. We will utilize the following important conclusion of the global minima manifold of $L(\cdot)$ \citep{lin2024exploring}:

\begin{lemma} 
Suppose that $m \geqslant M$. Let $S_0=\left\{(0, \ldots, 0) \in \mathbb{R}^d\right\}, S_j=\left\{\alpha e_j: \alpha > 0\right\}$ for $j \in[M]$, and $S=\cup_{j=0}^M S_j$. Then $\mathcal{M}$ is compact and can be analytically characterized as follows
\begin{align*}
    &\mathcal{M}=\left\{ \vphantom{\sum_{i=1}^m}\theta=\left(w_1, \ldots, w_m\right)\trans \in \mathbb{R}^{m \times d}: \forall i \in[m], w_i \in S \text { and } \forall j \in[M], \sum_{i=1}^m w_{i, j}=1\right\}.
\label{eq:manifold}
\end{align*}

\label{thm:minima}
\end{lemma}
This characterization of the  global minima manifold will play a critical role in our  analysis.
Note that $S_j \cap S_k=\emptyset$ for any $j \neq k \in\{0,1, \ldots, M\}$. Hence Lemma \ref{thm:minima} implies the following facts about the global minima:
\begin{itemize}
    \item There are at most $m+1$ types of student neurons, represented by $S_0, S_1, \ldots, S_M$, no matter how overparameterized the student network is. Moreover, for any $j \in[M]$, there exists at least one student neuron taking the type of $S_j$.
    \item For each neuron, there exists at most one coordinate to be nonzero and moreover, the coordinates from $M+1$ to $d$ must be zero.
\end{itemize}


\section{Overlap Analysis}\label{subsec:overlap}

In this section, we will conduct an intuitive but effective analysis on how the barrier between two global minima changes with the number of student neurons $m$ and teacher neurons $M$ using the overlap between two minima.

According to Lemma \ref{thm:minima}, we denote the global minima $\theta$ as a weight matrix $W \in \mathbb{R}^{m \times d}$, and each row of this weight matrix is a neuron: $W = (w_1,w_2,\cdots,w_m)\trans, w_i \in \mathbb{R}^d$. With previous characterization if the $k$-th component of a neuron is a non-zero element, then the row/ the neuron is said to belong to \textbf{Type k} , $k \in [M]$. We can also write $w_i \in S_j$ using the notation in Lemma \ref{thm:minima}.

We also denote $\alpha = (\alpha_1,\cdots,\alpha_M)$ to be the \textbf{type vector} of the global minima $W$. As there is a constraint that for $\forall j \in [M]$ there exists at least one student neuron in type $j$, we let $\alpha_j + 1$ be the total number of neurons belonging to type $j$, or the non-zero elements in the $k$-th column of $W$, $k \in [M]$. Then $\sum_{i=1}^M \alpha_i = m-M$. To further analyze the overlap between two global minima, we make the following Uniform Distribution assumption of our solution obtained by GD/SGD on the loss landscape.

\begin{assumption} (Uniform Distributed Solution)
    Each neuron is equally likely to be assigned to each type $S_j$,$j\in [M]$, that is, the type vector follows a multinomial distribution, $\alpha \sim \text{Multi}(m-M;\frac1M,\cdots,\frac1M)$. The actual number of neurons of type $j$ is $\alpha_j+1$. Moreover, the non-zero elements of neurons in any type follow a uniform distribution on the simplex.
\label{assu:uniform}
\end{assumption}

In the absence of other prior knowledge, this uniform distribution assumption is natural, as all neurons are created equal, and the probability of neurons being allocated to each type should be the same. And we also assume that $\forall i \in [m], w_i \notin S_0$, that is, there is no sparsity for neurons. We will talk about the sparsity of global minima in Section \ref{sec:sparsity}, and we also validated this uniform distribution assumption through some simulation experiments in Appendix \ref{subsec:uniform}. Here we set the parameter of multinomial distribution as $m-M$ to tackle the constraint that each type has at least one neuron. 

The following matrix gives an example of a weight matrix (\ref{eq:exmatrix}) on the global minima manifold $\mathcal{M}$. We say that two global minima $W\uo,W\ut$ \textbf{match}, as long as their type vectors $\alpha\uo,\alpha\ut$ coincide. In other words, each row in these two weight matrices is the same type. It's easy to see that if $W\uo, W\ut$ match, then for $\forall \lambda \in [0,1], \lambda W\uo + (1-\lambda)W\ut \in \mathcal{M}$, and the linear mode connectivity holds, which means the barrier between these two global minima is 0.

\begin{equation}
    W = 
\begin{blockarray}{ccccccccc}
 1 & 2 & 3 & \cdots & M & M+1 & \cdots & d & \\
\begin{block}{(cccccccc)c}
 0 & 1 & 0&\cdots & 0& 0 & \cdots & 0 & \ \text{Type 2}\\
 0 & 0 & \frac{1}{2}& \cdots & 0 & 0& \cdots & 0 &\ \text{Type 3}\\
 \vdots & \vdots & \vdots& \ddots & \vdots & \vdots& \ddots & \vdots \\
 0 & 0 &0 & \cdots & \frac{1}{3}& 0 & \cdots & 0 & \ \text{Type M}\\
 \vdots & \vdots & \vdots& \ddots & \vdots & \vdots& \ddots & \vdots \\
\frac{1}{2} & 0 & 0& \cdots & 0 & 0& \cdots & 0 & \ \text{Type 1}\\
\end{block}
\end{blockarray}
\label{eq:exmatrix}
\end{equation}

However, as the type of neurons is uniformly randomly assigned, there is little chance that two global minima on the manifold have the same type of vectors. Therefore, we will analyze the degree of matching using the ``\textbf{overlap}'' of two global minima $W\uo, W\ut$.

\begin{definition}
    Let $\alpha\uo,\alpha\ut$ be the type vectors of two solutions $W\uo,W\ut$, then the \textbf{overlap} between two type vectors $\alpha\uo,\alpha\ut$ (or two solutions) is defined as 
    
    \begin{equation}
        C(\alpha\uo,\alpha\ut):= \sumM \min(\alpha\uo_j,\alpha\ut_j)+M.
    \end{equation}
\end{definition}

A neuron of a solution overlapping means it can be matched with a certain neuron in the other solution, so this pair incurs no barrier. Those not overlapped can't be matched to any neuron in the other solution, incurring a high barrier. Therefore, the proportion of overlapping 
\begin{equation}
    P = C(\alpha\uo,\alpha\ut)/m
    \label{eq:proportion}
\end{equation}
indicates a good property of permutation invariance. For instance, if $P=1$, meaning two solutions can exactly match, then the barrier should be 0; on the other hand, if $P=0.5$, say, then half of the neurons cannot be matched, incurring a high barrier.

\begin{figure}[!htb]
\centering
\begin{minipage}[c]{0.45\textwidth}
  \includegraphics[width=\textwidth]{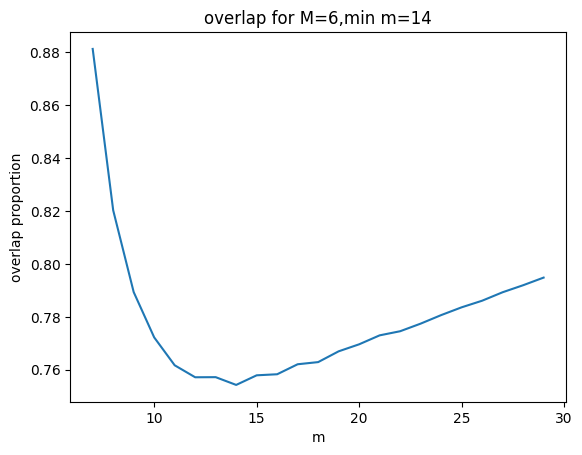}
  \subcaption{$M=6$}
\end{minipage}
\begin{minipage}[c]{0.45\textwidth}
  \includegraphics[width=\textwidth]{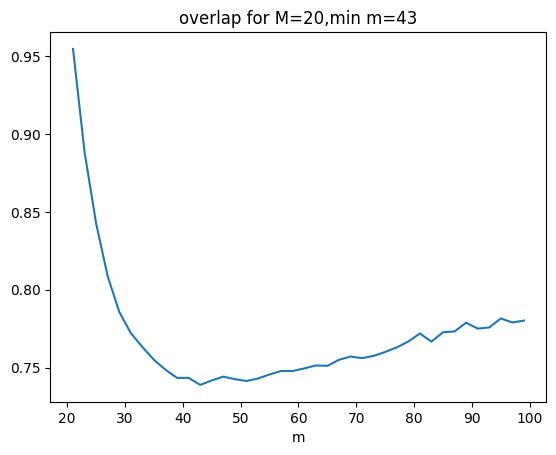}
  \subcaption{$M=20$}
\end{minipage}
\caption{The overlap proportion curve for different $M$. The left panel shows the overlap proportion $P$ for $M=6$ and the right panel shows that for $M=20$. Each data point is averaged over $10^4$ simulations.}
\label{fig:overlappro}
\end{figure}

Under the distributional assumption \ref{assu:uniform}, we can directly obtain how the overlap proportion of two solutions changes with the number of student neurons $m$:

\begin{theorem}
\label{thm:overlap}
    Assume two global minima $W\uo,W\ut$ follow the uniform distribution as in Assumption \ref{assu:uniform}, and we fix the number of teacher neurons $M$, then their expected overlap proportion $T(m,M) := \EE P = \dfrac{\EE C(\alpha\uo,\alpha\ut)}{m} \to 1$ when $m \to \infty$. Moreover, when $M \to \infty$, the limit value $\lim\limits_{m,M\to\infty}T(m,M)$ is minimized with $\lim\limits_{m,M\to \infty} \frac{m}{M}=2$.
\end{theorem}

Theorem 5 means that in a limiting sense, the expected overlap proportion will be minimized when $m=2M$. In Figure \ref{fig:overlappro} we generated multinomial type vectors and calculated their overlap proportion as defined in (\ref{eq:proportion}). We can observe that the overlap decreases first and then increases as $m$ increases, reaching its minimum value when $m$ is approximately equal to $2M$; As $m$ tends to infinity the proportion approaches 1. This result is consistent with the previous Theorem \ref{thm:overlap}.

\section{Barrier Calculation}

When we consider the overlap of two global minima, we ignore that the non-zero elements also decay as $m$ increases. If the non-overlap neurons have very small norms, the barrier will be very low even if the overlap proportion is away from 1. So we will employ a refined analysis on the barrier in this subsection.

Firstly, we need to figure out what kind of permutation we should adopt for two given solutions, in order to minimize the barrier as much as possible. The following algorithm tells us how to find such a permutation.

\begin{algorithm}
\caption{Find the best permutation for global minima $W\uo,W\ut$}
\label{alg:permutation}
\KwData{Two global minima $W\uo$ and $W\ut$}
\KwResult{Permuted $\widetilde{W}\ut$}
Sort non-zero elements of each column of $W\uo,W\ut$ in descending order\;
Calculate the index set $I\uo_j$ for $W_1$ ($j \in [M]$)\;
num[$j$]=1, for $j\in[M]$\;
\For{$i = 1$ \KwTo $m$}{
  \For{$j = 1$ \KwTo $M$}{
   \If{$w\ut_i \in S_j$ and num[$j$] $\leqslant |I\uo_j|$}{
    index = $I\uo_j$[num[$j$]]\;
    $\widetilde{W}\ut[\text{index},] = w\ut_i$\;
    num[$j$] +=1\;
   }
  }
}
Fill the unassigned neurons of $W\ut$ sequentially into the empty rows of $\widetilde{W}\ut$\;
\Return{$\widetilde{W}\ut$}\;
\end{algorithm}

In the Algorithm \ref{alg:permutation}, we try to find the best permutation of the neurons of $W\ut$ so that the barrier of linear interpolation of $W\uo$ and permuted $\widetilde{W}\ut$ will be low. For each student neuron $w_i\ut \in S_j$, we assign it to the corresponding type $S_j$ of $W\uo$ until all the neurons of $j$-th type in $W\uo$ are already matched. Moreover, we sort the non-zero elements of $W\uo,W\ut$ first, so the neuron with a large norm will be matched first and the rest unassigned neurons of $W\ut$ have a small norm.

In order to theoretically calculate the barrier after the best permutation, we use the kernel methods to analyze the loss function (\ref{eq:lossfun}). Assume that $\pi=\rho=\tau_{d-1}$. By the rotational invariance, $k_\pi$ can be written in a dot-product form:
\begin{equation}
    k_\pi\left(x, x^{\prime}\right)=\int_{\mathbb{S}^{d-1}} \sigma\left(v\trans x\right) \sigma\left(v\trans x^{\prime}\right) \mathrm{d} \tau_{d-1}(v)=\kappa\left(x\trans x^{\prime}\right),
\end{equation}
where $\kappa : [-1,1] \to \mathbb{R}$. And for ReLU activation, we have \citep{cho2009kernel,leiwu22spectral}
\begin{equation}
    \kappa(t)=  \frac{1}{2 \pi d}\left((\pi-\arccos t) t+\sqrt{1-t^2}\right).
\end{equation}

From our loss function (\ref{eq:lossfun}), we have the following derivation:
\begin{align*}
    L(W) = & \mathbb{E}_{x \sim \tau_{d-1}}\left[\left(\sum_{i=1}^m \sigma\left(w_i\trans x\right)-\sum_{j=1}^M \sigma\left(e_j\trans x\right)\right)^2\right] \\
    = & \EE \left[(\sum_{i=1}^m \sigma\left(w_i\trans x\right))^2+(\sum_{j=1}^M \sigma\left(e_j\trans x\right)^2) -2(\sum_{j=1}^M \sigma\left(e_j\trans x\right))(\sum_{i=1}^m \sigma\left(w_i\trans x\right))\right]
    \\
    = & \sum_{i,i'} |w_i||w_{i'}|\kappa(\Tilde{w_i}\trans \Tilde{w_{i'}}) + \sum_{j,j'} \kappa(e_j\trans e_{j'}) - 2 \sum_{i,j} |w_i|\kappa (\tilde{w_i}\trans e_j),
\end{align*}
where $\Tilde{w_i}$ means the normalized vector with $L_2$ norm 1. With this characterization of the loss function, we have the following theoretical description of the barrier curve:

\begin{theorem}
\label{thm:decay}
    For any two solutions $W\uo,W\ut$ following the Assumption \ref{assu:uniform}, denote $\widetilde{W}\ut$ as the permuted solution obtained by the Algorithm \ref{alg:permutation}. The barrier is defined as
    \begin{equation}
    \begin{split}
     &B(W\uo,W\ut) := \sup_{\lambda \in [0,1]} \{ L(\lambda W\uo + (1-\lambda) \widetilde{W}\ut) - \lambda L(W\uo) - (1-\lambda) (W\ut)  \}.
    \end{split}
    \end{equation}
    Then we have $B(W\uo,W\ut) \to 0$ when $m \to \infty$, and the decay rate is  $B(W\uo,W\ut) = O(m^{-1/2})\ (m \to \infty)$.
\end{theorem}

In order to characterize this rate of decrease in the barrier curve when $m\to \infty$, we need to assume the following approximation: for each type of neurons corresponding to every solution, the proportion of matching neurons is uniformly $\gamma$. The detailed meaning is as follows: Let $W\uo,W\ut$ be two solutions, then every neuron $w\uo_i,w\ut_i \in S_j$ ($j\in [M]$) as in Theorem \ref{thm:minima}. For simplicity, we just assume $W\ut$ is already permuted by the Algorithm \ref{alg:permutation}. Let $I\uk_j = \{i: w\uk_i \in S_j\}, j\in[M], k=1,2, I_j = I\uo_j \cap I\ut_j$. Then our assumption is
\begin{equation}
    \sum_{i \in I_j} |w\uk_i| = \gamma,\quad \forall j \in [M],k=1,2. 
\end{equation}
The detailed proof can be found in Appendix \ref{appen:pf3}.

\begin{figure}[!htb]
    \centering
    \includegraphics[width = 0.55\textwidth]{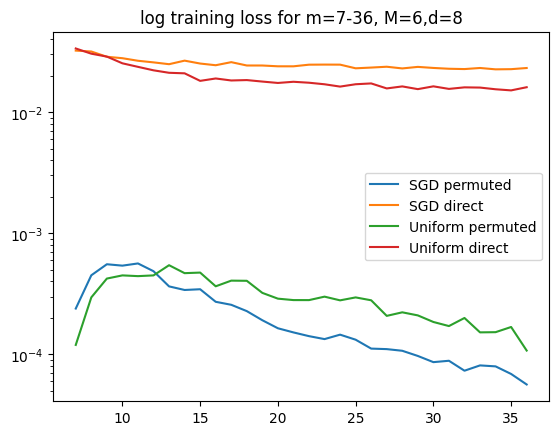}
    \caption{The log barrier curve for SGD solutions and uniformly sampled solutions. The number of teacher neurons $M = 6$, dimension is $d = 8$, and the number of student neurons $m$ is varied from 7 to 36. Each data point is an average of 20 independent realizations.}
    \label{fig:barrier1}
\end{figure}

We conducted simulation for the barrier of direct linear interpolation and the barrier after the best permutation. We also considered the minima obtained by SGD and the minima uniformly sampled from the manifold $\mathcal{M}$. Some detailed simulation settings can be found in Appendix \ref{subsec:detail}.

Figure \ref{fig:barrier1} displays the changing trend of the barrier between two global minima found by SGD and those found by the uniform distribution on the manifold as \(m\) varies. We plot the barriers directly connecting the two global minima found by GD and those connecting after finding the optimal permutation using the aforementioned Algorithm \ref{alg:permutation}. On the one hand, the barrier significantly decreases after the optimal permutation, demonstrating the correctness of permutation invariance. On the other hand, we can also observe that the barrier after permutation increases first and reaches its maximum when \(m=2M\); after that it decreases to 0 as \(m\) increases to infinity. This barrier curve obtained from simulation on the two-layer teacher-student ReLU network validates both the overlap Theorem \ref{thm:overlap} and the barrier Theorem \ref{thm:decay}. It's also worth noting that when student neurons $m$ is relatively large, the barrier of permuted SGD solutions is slightly smaller than the barrier of permuted uniformly sampled solutions. This is partly due to the sparsity of GD solutions when $m$ is large and the learning rate is high. We will further discuss the sparsity of global minima in the following Section \ref{sec:sparsity}. 

\begin{figure}[!htb]
    \centering
    \includegraphics[width = 0.55\textwidth]{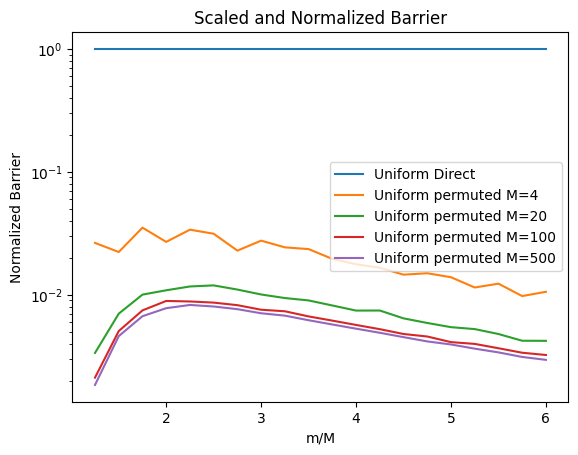}
    \caption{The normalized log barrier curve for uniformly sampled solutions. The barrier for direct linear interpolation in each setting with different $M$ is normalized to 1, and we plot the relative barrier for permuted solutions with different numbers of teacher neurons $M = 4,20,100,500$. $x$-axis is $m/M$ and $y$-axis represents normalized barrier = $\text{Barrier}_{\text{Permuted}}/\text{Barrier}_{\text{Direct}} $. Each data point is an average of 50 independent realizations.}
    \label{fig:barrier2}
\end{figure}

These phenomena are more evident in Figure \ref{fig:barrier2}, where we have normalized the barrier concerning the direct linear interpolation. We only use uniform samples in order to alleviate computational costs here, because the behavior of SGD solutions and uniformly sampled solutions are very similar, as can be seen in Figure \ref{fig:barrier1}. We can observe that the optimal permutation can reduce the barrier by $10^{-2}$, and the phenomenon of the barrier reaching its maximum when \(m=2M\) becomes more evident as the number \(M\) of teacher neurons increases. Permutation also brings more benefits when $M$ is larger. The barrier curve for more settings can be found in Appendix \ref{subsec:morebarrier}.

\begin{figure}[!htb]
    \centering
    \includegraphics[width = 0.6\textwidth]{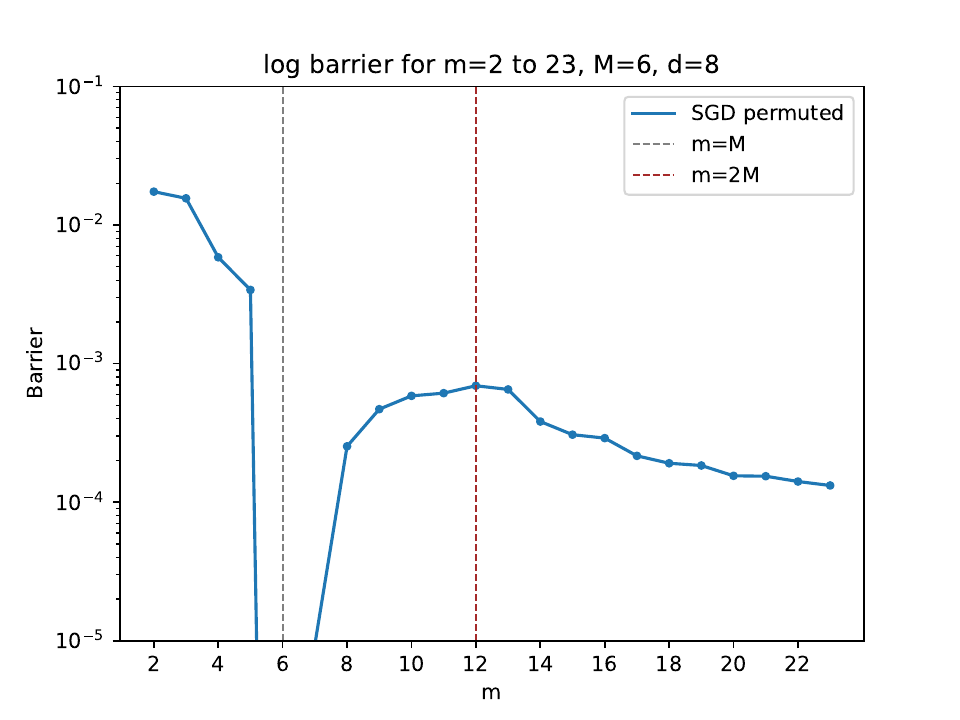}
    \caption{ {\bf The double descent phenomenon for LMC modulo permutation.}
    Barrier as a function of the number of student neurons \(m\). The first descent appears as \(m\) approaches \(M\) (under-realization regime), and the second descent occurs as \(m\) exceeds \(2M\), illustrating the ``double descent'' phenomenon. Note that when $m=M$, the student neurons can always match teachers and thus barrier is 0.}
    \label{fig:doubledescent}
\end{figure}

\paragraph*{Double descent.}
To further investigate the interplay between network size and loss barriers, we extend our analysis to the under-realization regime $(m<M)$ and observe a clear ``double descent'' phenomenon in Figure \ref{fig:doubledescent}. Specifically, the first descent appears when \(m\) increases toward \(M\). In this under-realization regime, each student solution can only align a subset of its neurons with those of the teacher, so two independently trained models tend to match different subsets of teacher neurons. This partial overlap leads to a non-trivial loss barrier between solutions that cannot be fully removed via permutation. As \(m\) approaches \(M\), the extent of “unmatched” teacher neurons in each solution decreases, thereby lowering the barrier. The second descent occurs once \(m\) exceeds approximately \(2M\), transitioning the student network into a regime where it has sufficient capacity to effectively match, and possibly surpass, the teacher neurons. The experimental results confirm both descents in the barrier size, demonstrating that the shift from under- to over-realization is key to understanding how network capacity influences solution alignment and, consequently, the loss landscape.

\section{Sparsity of Global Minima}
\label{sec:sparsity}
In previous discussions, we have been assuming under Assumption \ref{assu:uniform} that the solutions found by GD and SGD satisfy the property of uniform distribution. This assumption holds true when the learning rate of GD and SGD is relatively small. Some validation can be found in Appendix \ref{subsec:uniform}. 

However, in actual experiments we implemented, when the learning rate is large, we observe that GD/SGD tends to find sparser solutions, which means there are neurons whose elements are all zero.

To model the sparsity of a weight matrix $W$, we use the \textit{PQ Index} (PQI) as a measure of sparsity. PQ Index describes the sparsity of a vector using the $L_p$ norm, which is first proposed in \cite{diao2023pruning}.
\begin{definition}
    For any $0<p<q$, the PQI of a non-zero vector $w \in \mathbb{R}^d$ is
\begin{equation}
    \mathbf{I}_{p, q}(w)=1-d^{\frac{1}{q}-\frac{1}{p}} \frac{\|w\|_p}{\|w\|_q},
\end{equation}
where $\|w\|_p=\left(\sum_{i=1}^d\left|w_i\right|^p\right)^{1 / p}$ is the $\ell_p$-norm of $w$ for any $p>0$. 
\end{definition}

The larger the PQI is, the more sparse the vector will be. For example, for a unit vector $e_1 = (1,0,0,\cdots,0)$, which is very sparse, we have $\mathbf{I}_{p, q}(e_1) = 1-d^{\frac{1}{q}-\frac{1}{p}}$; for a uniform vector $w = (1,1,\cdots,1)$, we have $\mathbf{I}_{p, q}(w) = 0$. For our global minima, we have two different ways to evaluate its sparsity. On the one hand, we can directly flatten the matrix into a vector and then calculate the PQI of this vector. On the other hand, due to the properties of the solution, each neuron actually only has one non-zero element, so we can also take the norm by row first, and then calculate the PQI of this norm vector. In all of our simulations, we set $p=0.5,q=1$ in the PQI.

\begin{figure}[!htb]
    \centering
    \includegraphics[width = 0.6\textwidth]{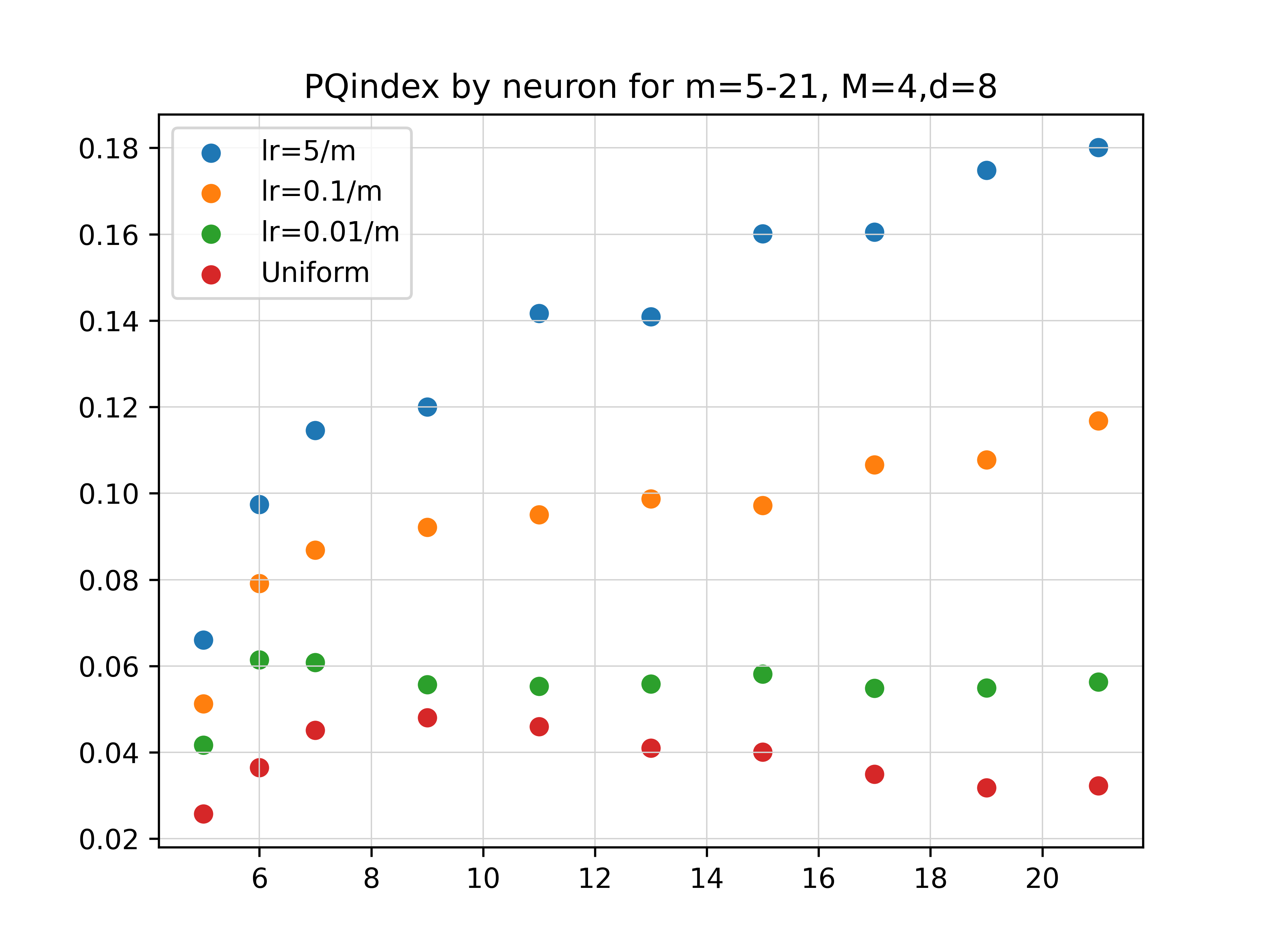}
    \caption{The PQI for the GD solution with different learning rate. $x$-axis is student neurons $m$ and $y$-axis is PQI by neuron. \textbf{Uniform} means we uniformly generate samples from the manifold. Each data point is an average of 100 independent realizations.}
    \label{fig:pqindex1}
\end{figure}

The sparsity of a solution obtained by GD/SGD is closely related to the learning rate, as indicated in the following experiments.
In Figure \ref{fig:pqindex1}, we plot the PQI of solutions obtained from GD with different learning rates. Note that we use learning rate that decays with the width $m$. We can clearly observe that with a smaller learning rate the PQI of minima is smaller, and the uniformly sampled minima have the smallest PQI or sparsity. When the learning rate is large, the PQI also increases with the increase in network width $m$, while this phenomenon is not evident when the learning rate is low.

As in the previous theoretical analysis, we assume that GD/SGD solutions follow a uniform distribution as Assumption \ref{assu:uniform} and there is no zero row in $W$. Although a high learning rate or large $m$ will encourage sparsity in the global minima, this sparsity is indeed beneficial for our desired permutation invariance, as those zero rows or zero neurons will incur no barrier when pairing, and the essential neurons causing overlap or barrier is smaller than total neuron number $m$. Therefore, in Figure \ref{fig:barrier1} the barrier of permuted SGD solutions is lower than that of permuted uniformly sampled solutions, and we attribute its reason to the sparsity of the SGD solutions.

\begin{figure}[!htb]
    \centering
    \includegraphics[width = 0.6\textwidth]{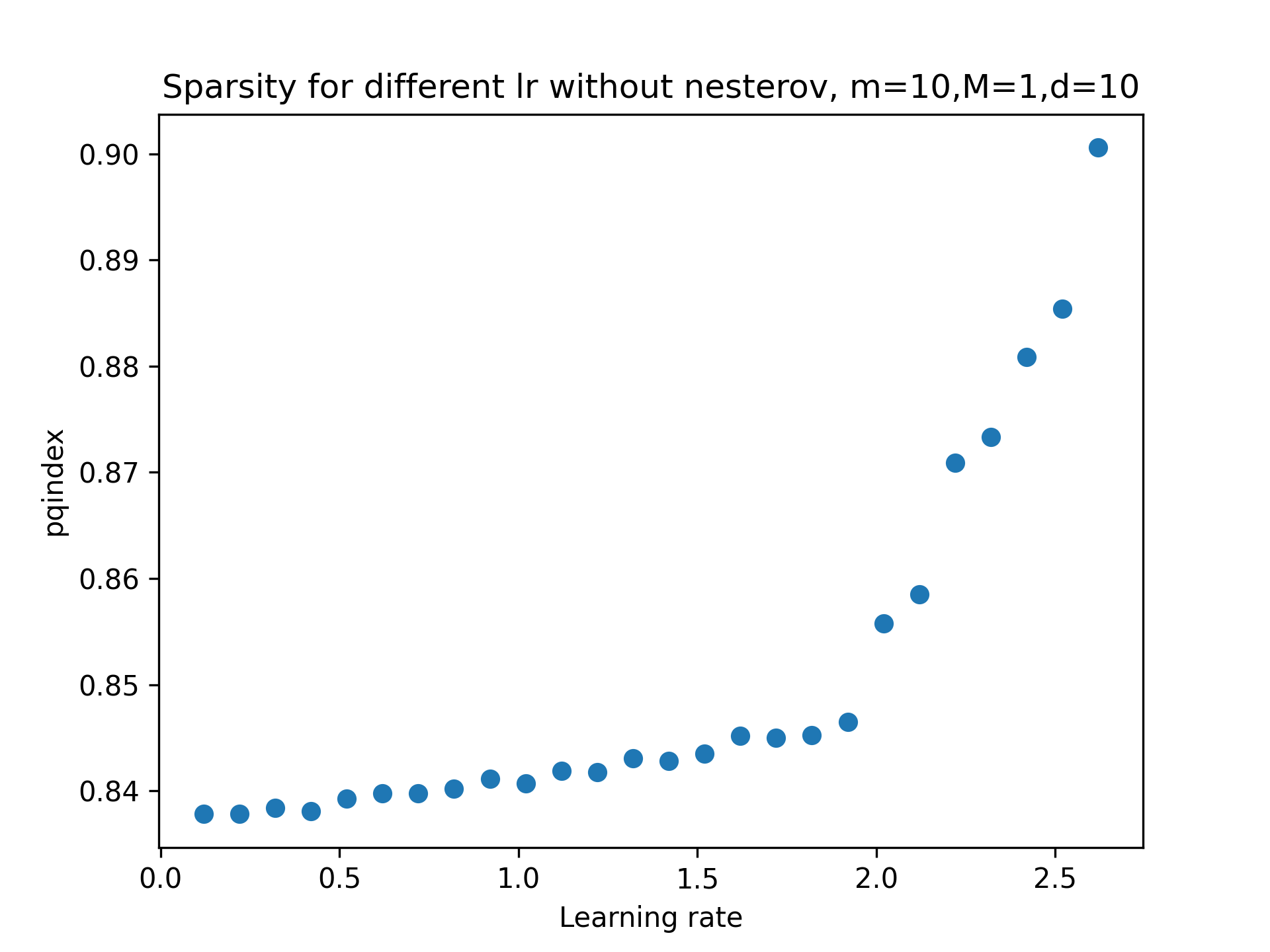}
    \caption{PQI of flattened $W$ with different learning rates. We trained the network using GD without Nesterov, setting a grid for different learning rates from 0.1 to 2.0. $y$-axis is the PQ Index of flattened parameter matrix $W$. }
    \label{fig:pqindex2}
\end{figure}

When we run gradient descent with different learning rates, we also observe a phase transition phenomenon of the sparsity of the solution. In Figure \ref{fig:pqindex2}, we plotted the PQI of the solution obtained by GD without Nesterov with different learning rates. When the learning rate is below a certain threshold, it has little effect on sparsity, which manifests in the specific solution as having no zero rows and slowly increasing PQI. However, when the learning rate is increased beyond this threshold, for example, learning rate $=2$ here, the resulting solution becomes sparse, zero rows appear in the solution, and the PQI also increases rapidly. This may be related to the loss landscape around global minima, and we look forward to future exploration providing a more detailed explanation of the threshold phenomenon in this setting.

\section{Empirical Investigations}
In this section, we empirically validate our theoretical findings beyond simulation data using two-layer teacher-student ReLU networks. For more complex architectures, such as multi-layer fully connected networks and CNNs, we apply the algorithms in \cite{benzing2022random,ainsworth2022git} to locate the best approximate permutation.

\begin{figure}[!htb]
    \centering
    \includegraphics[width=0.7\linewidth]{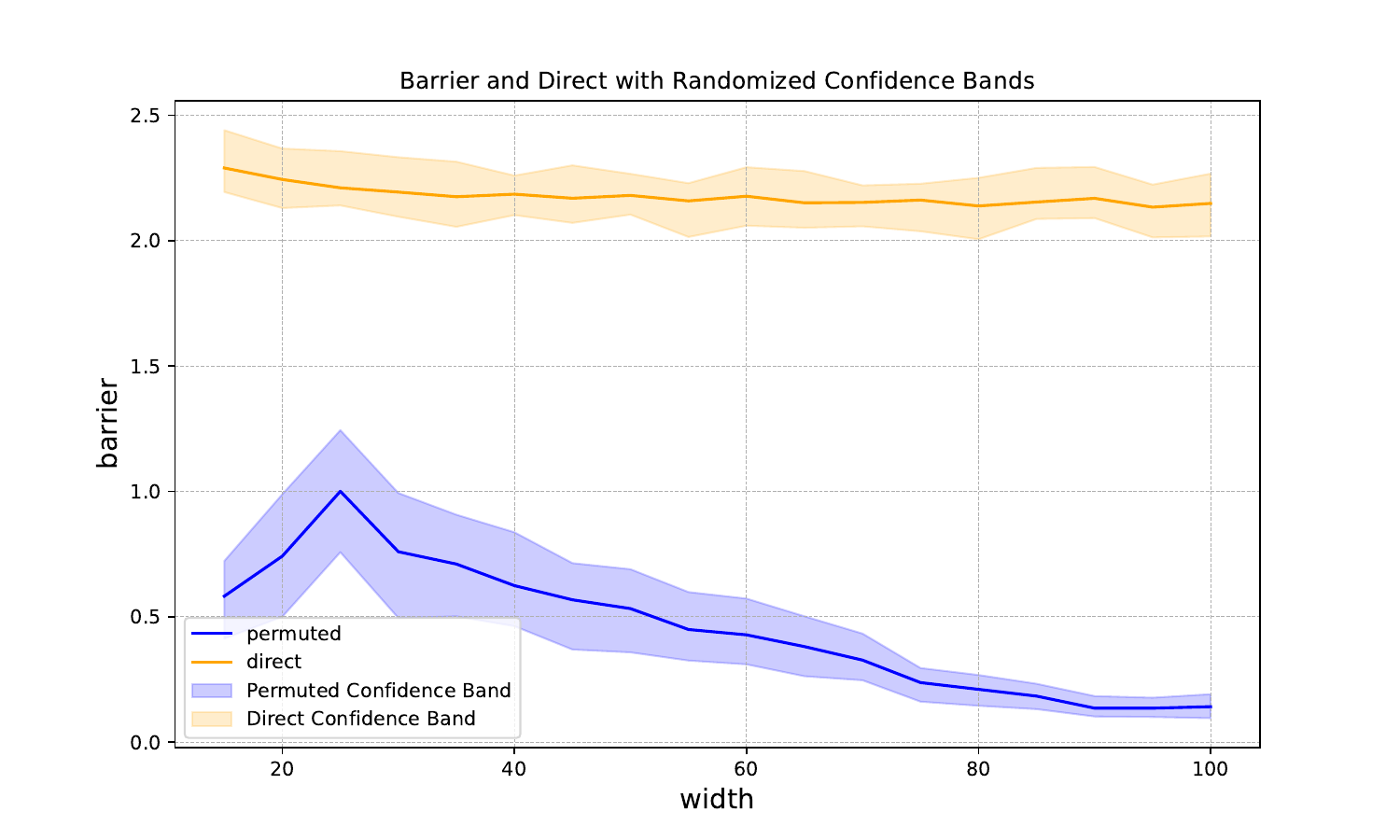}
    \caption{Barrier under different widths of the 4-layer MLP trained on MNIST with bands of top 90\% and 10\% percentile, for permuted and direct interpolation, respectively. Each result is an average of 10 independent realizations.}
    \label{barrier_mnist}
\end{figure}

We first train  4-layer fully-connected neural networks for fitting the MNIST dataset, with a learning rate of 0.05. Figure \ref{barrier_mnist} shows the LMC barrier (the negative log-likelihood) modulo permutation under different model widths from 15 to 100. It is clear that the barrier goes up and then goes down as the width increases, exhibiting a peak phenomenon. This  is aligned with our theoretical analysis and previous simulation results.

\begin{figure}[!htb] 
\minipage{0.33\textwidth}
  \includegraphics[width=\linewidth]{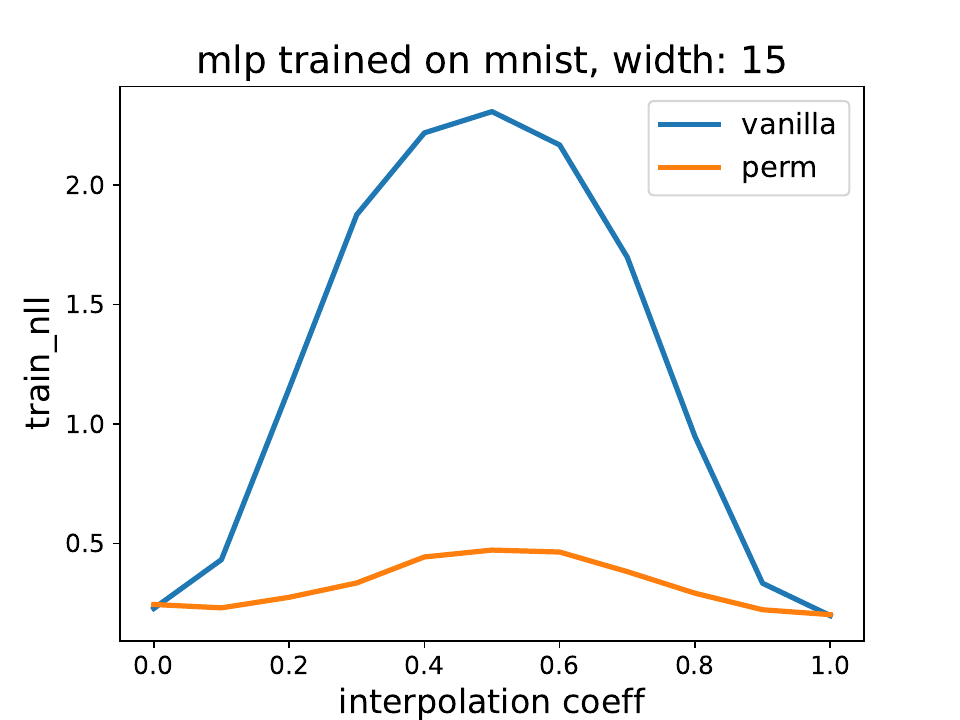}
\endminipage\hfill
\minipage{0.33\textwidth}
  \includegraphics[width=\linewidth]{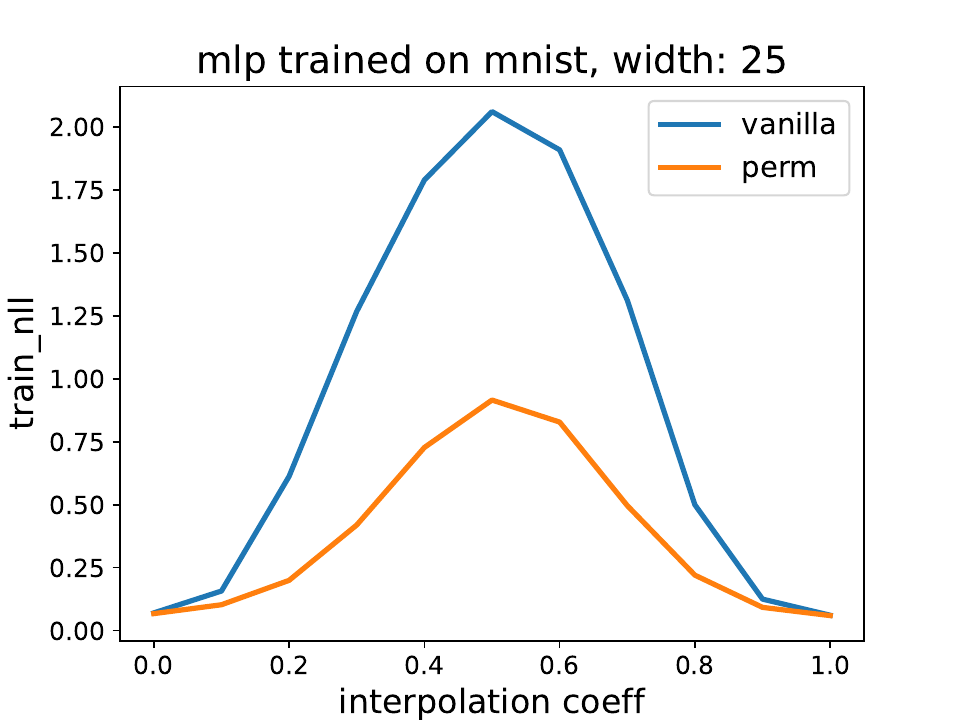}
\endminipage\hfill
\minipage{0.33\textwidth}%
  \includegraphics[width=\linewidth]{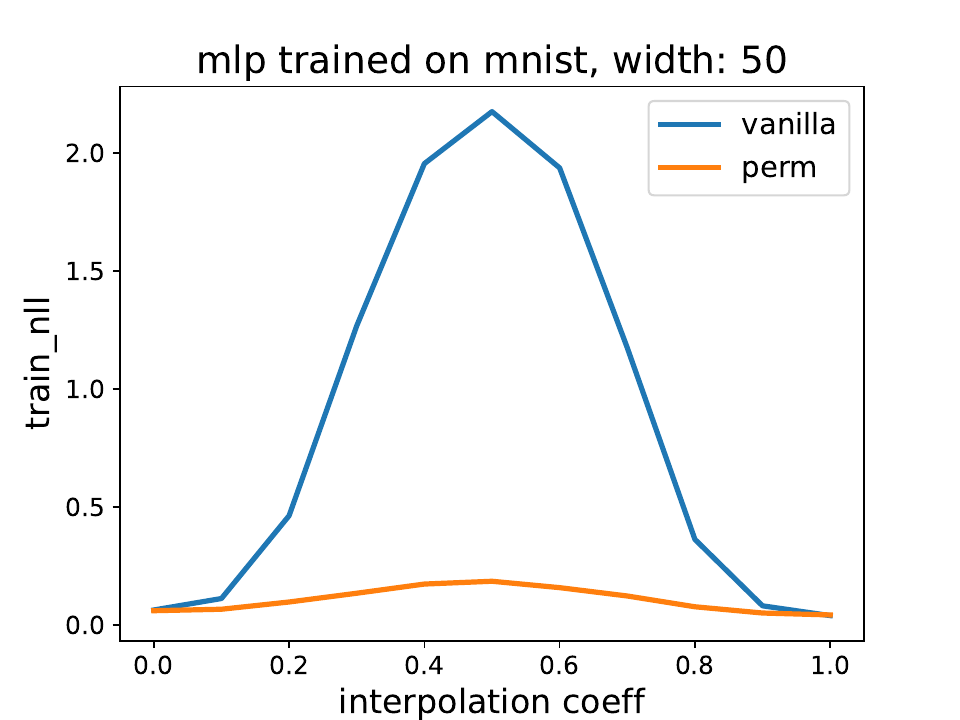}
\endminipage
\caption{Interpolation NLL plot under different widths.}
\label{interpolation}
\end{figure}

Figure \ref{interpolation} further shows the interpolation  plot comparing the loss on the line connecting original models and the line connecting permuted models. As we can see, the role of permutation invariance depends  greatly on the model width. When the model width is $m=25$, the effect of permutation is pretty limited, which is consistent with the peak value in Figure \ref{barrier_mnist}. For direct linear interpolation, the NLL is constantly large for all widths we have examined.

\section{Conclusion and Discussion}
In this paper, we analyzed the role of permutation invariance in a two-layer ReLU network under a teacher-student regime from a theoretical perspective. We showed that as network width increases, the barrier of the linear connecting path between the permuted minima has a trend of first increasing and then decreasing to 0, with the maximum value at $m=2M$. Further, we found that GD/SGD solutions have an increasing sparsity in learning rate with a phase transition pattern. Sparsity is beneficial to permutation invariance, hence this phenomenon serves as a complement to our theoretical analysis where we assume uniform distribution over global minima manifold. We empirically verified our results by conducting experiments on simulation data and MNIST datasets. The results explained why permutation invariance would appear significant or negligible under different conditions.  

For future work, it remains an open problem why increasing learning rates yields sparser GD/SGD solutions.  The role of permutation invariance with model depth is also a problem worth working on in the future. It is an intriguing question how the peak value of permuted barrier and the gap between permuted barrier and direct barrier change with width, depth, and network structure in various neural networks.

\section*{Acknowledgments}
Lei Wu is supported by the National Key R\&D Program of China (No.~2022YFA1008200) and National Natural Science Foundation of China (No.~12288101). We sincerely appreciate the constructive feedback from the anonymous reviewers.

\addcontentsline{toc}{section}{References}
\bibliographystyle{apalike}
\bibliography{ref}

\newpage
\addcontentsline{toc}{section}{Appendix}

\appendix

\section{Proofs}

\subsection{The Proof of Theorem \ref{thm:overlap}}

Based on the assumptions we have made, the marginal distribution of $\alpha_j$ is Bi$(m-M,1/M)$, subject to the constraint that $\sumM \alpha_j = m-M$. The $\alpha_j$'s are identically distributed, though not independent. $\alpha\uo,\alpha\ut$ are independent. Then with the linearity of expectations, we have 
\begin{equation}
    T(m,M) := \frac{\sumM \EE \min(\alpha\uo_j,\alpha\ut_j)+M}{m} = \frac{M}{m}(\EE \min(X,Y)+1)
\end{equation}
with $X,Y$ i.i.d $\sim Bi(m-M,\frac{1}{M}).$


Equivalently we show that as a function of $m$, $T(m,M)$ from $m = M$ initially increases monotonically and then decreases monotonically. Here,
\begin{equation}
    T(m,M)  = \frac{M}{m}\left(\frac{m-M}{M}+1 - \dfrac{1}{2}\EE |X-Y|\right) = 1 - \frac{M}{2m}\EE |X-Y|.
\end{equation}

So we have the limit result. $(\EE|X-Y|)^2 \leqslant \EE(X-Y)^2 = 2 \Var X = \dfrac{2(m-M)(M-1)}{M^2}$. Hence $T(m,M) \geqslant 1- \dfrac{\sqrt{2(m-M)(M-1)}}{2m} \to 1$ when $m \to +\infty$.

With $X,Y$ i.i.d $\sim Bi(m-M,\frac{1}{M}),$ and $m,M \to \infty$, we have the following central limit theorem:
\begin{equation}
    \frac{X-\frac{m-M}{M}}{\sqrt{(m-M)\frac{M-1}{M^2}}} \to_d N(0,1); \quad\frac{Y-\frac{m-M}{M}}{\sqrt{(m-M)\frac{M-1}{M^2}}} \to_d N(0,1)
\end{equation}

For $\xi_1,\xi_2 \sim N(0,1)$ i.i.d., $\EE |\xi_1-\xi_2| = \frac{2}{\sqrt{\pi}}$. Therefore 
\begin{equation}
    \lim_{m,M\to \infty} T(m,M) = 1- \lim_{m,M\to \infty}\frac{M}{2m}\sqrt{\frac{(m-M)(M-1)}{M^2}} \frac{2}{\sqrt\pi}.
\end{equation}
Let $t=\lim_{m,M\to \infty}\frac{M}{m}$, then we have
\begin{equation}
    \lim_{m,M\to \infty}T(m,M)  =  1 - \sqrt{\frac{t(1-t)}{\pi}}.
\end{equation}
Then in this limiting sense, the overlap is minimized when $t=\frac12$, which is $m=2M$. \qed

\subsection{The Proof of Theorem \ref{thm:decay}} \label{appen:pf3}
We first derive the loss function as
\begin{align*}
    L(W)=& \mathbb{E}_{\mathbf{x} \sim \tau_{d-1}}\left[\left(\sum_{i=1}^m \sigma\left(w_i\trans x\right)-\sum_{j=1}^M \sigma\left(e_j\trans x\right)\right)^2\right] \\
    = & \EE [(\sum_{i=1}^m \sigma\left(w_i\trans x\right))^2+(\sum_{j=1}^M \sigma\left(e_j\trans x\right)^2)-2(\sum_{j=1}^M \sigma\left(e_j\trans x\right))(\sum_{i=1}^m \sigma\left(w_i\trans x\right))] \\
    = & \sum_{i,i'} |w_i||w_{i'}|\kappa(\Tilde{w_i}\trans \Tilde{w_{i'}}) + \sum_{j,j'} \kappa(e_j\trans e_{j'}) - 2 \sum_{i,j} |w_i|\kappa (\tilde{w_i}\trans e_j)  \\
    =: & L_1 + L_2 - 2L_3,
\end{align*}

where $\Tilde{w_i}$ means the normalized vector with $L_2$ norm 1. The simplest term is $\sum\limits_{j,j'} \kappa(e_j\trans e_{j'})$. For $j=j'$, the inner product is 1; for $j\neq j'$, the inner product is 0. So we have $L_2 = \sum\limits_{j,j'} \kappa(e_j\trans e_{j'}) = M\kappa(1) + (M^2-M) \kappa(0)$.

We further assume that $I_c = \bigcup_{j=1}^M I_j, I_r = [m] \backslash I_c$. Now for an interpolation after best permutation, we permute the neurons of $W\ut$ so that all neurons that can be matched are matched. We denote that $W := \lambda W\uo + (1-\lambda) \widetilde{W}\ut$, where $\widetilde{W}\ut$ means the solution after an appropriate permutation. Then we know that for $i \in I_j$, the neurons of interpolation also satisfy that $w_i \in S_j, \forall j \in [M]$. But for $i \in I_r$, the neuron $w_i$ will have two non-zero elements, thus not belonging to any $S_j$.

Then we can write $L_1$ as
\begin{align*}
    L_1 = & \sum_{i,i'}^m |w_i||w_{i'}|\kappa(\Tilde{w_i}\trans \Tilde{w_{i'}}) \\
     = & \sum_{i,i'\in I_c} |w_i||w_{i'}|\kappa(\Tilde{w_i}\trans \Tilde{w_{i'}}) + \sum_{i,i'\in I_r} |w_i||w_{i'}|\kappa(\Tilde{w_i}\trans \Tilde{w_{i'}}) + 2\sum_{i\in I_r,i'\in I_c} |w_i||w_{i'}|\kappa(\Tilde{w_i}\trans \Tilde{w_{i'}})  \\
     =: & L_{11} + L_{12} + 2L_{13}. \\
    L_{11} = & \sum_{j=1}^M \sum_{i \in I_j} \left[ \sum_{i' \in I_j} |w_i||w_{i'}|\kappa(\Tilde{w_i}\trans \Tilde{w_{i'}}) + \sum_{i' \in I_{j'},j'\neq j} |w_i||w_{i'}|\kappa(\Tilde{w_i}\trans \Tilde{w_{i'}})\right] \\
    = &  \sum_{j=1}^M \sum_{i \in I_j} \left[ \sum_{i' \in I_j} |w_i||w_{i'}|\kappa(1) + \sum_{i' \in I_{j'},j'\neq j} |w_i||w_{i'}|\kappa(0) \right].
\end{align*}

For $w_i, i \in I_c$, $w_i \in S_j$ for some $j \in [M]$, so the norm of $w_i$ is the value of its non-zero element. So we can derive that 
\begin{align*}
    L_{11} = & \sum_{j=1}^M \left[ \sum_{i \in I_j}  \sum_{i' \in I_j} |w_i||w_{i'}|\kappa(1) + \sum_{i' \in I_{j'},j'\neq j} |w_i||w_{i'}|\kappa(0) \right] \\
    = & \sum_{j=1}^M [\gamma^2 \kappa(1)+(M-1)\gamma^2
    \kappa(0)] = M\gamma^2 \kappa(1)+M(M-1)\gamma^2
    \kappa(0).
\end{align*}
We can have the following estimation of the non-matching part $I_r$:
\begin{align}
    \sum_{i\in I_r} |w_i| = \sum_{i \in I_r} \sqrt{\sum_{j\in [M]} w_{ij}^2} \leqslant & \sum_{i \in I_r} \sum_{j\in [M]} w_{ij} = M(1-\gamma) \\
    \geqslant & \frac{\sqrt{2}}{2} \sum_{i \in I_r} \sum_{j\in [M]} w_{ij} = \frac{\sqrt{2}}{2}M(1-\gamma) .
\end{align}
because there are only two non-zero elements in each $w_i, i \in I_r$, so here the coefficient is $\dfrac{\sqrt2}{2}$. Then $L_{12}$ and $L_{13}$ can be estimated:
\begin{align*}
    L_{12} &\leqslant \sum_{i,i'\in I_r} |w_i||w_{i'}| \kappa(1) \leqslant M^2 (1-\gamma)^2 \kappa(1), \\
    L_{13} & =  \sum_{i\in I_r} \sum_{i'\in I_c} |w_i||w_{i'}|\kappa(\Tilde{w_i}\trans \Tilde{w_{i'}}) 
    \leqslant M\gamma(1-\gamma) \kappa(1).
\end{align*}

Similarly, we can estimate the lower bound of $L_{3}$:
\begin{align*}
    L_3 = & \sum_{i,j} |w_i|\kappa (\tilde{w_i}\trans e_j) 
    = \sum_{i \in I_c,j} |w_i|\kappa (\tilde{w_i}\trans e_j) + \sum_{i\in I_r,j} |w_i|\kappa (\tilde{w_i}\trans e_j) \\
    = & \sum_{j'=1}^M \sum_{i \in I_{j'}} |w_i| \left(\kappa(\tilde{w_i}\trans e_{j'}) + \sum_{j\neq j'}\kappa(\tilde{w_i}\trans e_j) \right)+\sum_{i\in I_r,j} |w_i|\kappa (\tilde{w_i}\trans e_j)\\
    = & M\gamma \kappa(1) + M(M-1) \gamma \kappa(0) +\sum_{i\in I_r,j} |w_i|\kappa (\tilde{w_i}\trans e_j) \\
    \geqslant & M\gamma \kappa(1) + M(M-1) \gamma \kappa(0) + \frac{\sqrt2}{2} M^2 (1-\gamma)\kappa(0).
\end{align*}

In conclusion, we have an upper bound for the entire loss:
\begin{align*}
    L = & L_1+L_2 -2L_3 \\
    \leqslant &  M\gamma^2 \kappa(1)+M(M-1)\gamma^2
    \kappa(0) + M^2 (1-\gamma)^2 \kappa(1) + M\gamma(1-\gamma) \kappa(1) + M\kappa(1) + (M^2-M) \kappa(0) \\
    & -2M\gamma \kappa(1) - 2M(M-1) \gamma \kappa(0) - \sqrt2M^2 (1-\gamma)\kappa(0) \\
    = & (1-\gamma)^2 \left[(M^2+M)\kappa(1) + (M^2-M)\kappa(0)\right] + (1-\gamma) \left[M\gamma\kappa(1)-\sqrt2M^2\kappa(0)\right] \\
    = & O(1-\gamma) = O(m^{-1/2}).
\end{align*}
\qed

\section{Additional Experiment Results}

\subsection{Detailed Setting for Experiments}
\label{subsec:detail}
For two-layer ReLU network in the teacher-student regime, we conduct gradient descent without nesterov. We use random initialization, where each element is a Gaussian noise with standard error $1/md$. For the uniform distribution on the manifold, we utilize the following characterization:
\begin{lemma}
    Let $\left(X_1, \ldots, X_n\right)$ be a random point uniformly distributed on the simplex\\ $\left\{\left(x_1, \ldots, x_n\right) \mid \sum_{k=1}^n x_k=1\right\}$. Then 
$$
\left(X_1, \ldots, X_n\right) \stackrel{d}{=} \frac{\left(Z_1, \ldots, Z_n\right)}{Z_1+\cdots+Z_n},
$$
 where $Z_1, \ldots, Z_n$ are i.i.d $\operatorname{Exp}(1)$ random variables. So, each $X_i$ equals
$$
\frac{Z_1}{Z_1+\cdots+Z_n}=\frac{Z_1}{n} / \frac{Z_1+\cdots+Z_n}{n}
$$
in distribution. Also, $\frac{Z_1+\cdots+Z_n}{n} \rightarrow 1$ almost surely and hence in distribution (as $n \rightarrow \infty$ ), by the strong law of large numbers. Thus, for each $i$, the distribution of $n X_i$ (not of $X_i$ ) goes to $\operatorname{Exp}(1)$.
\label{lemma:exp}
\end{lemma}
Therefore, we first generate the type vector $\alpha = (\alpha_1,\cdots,\alpha_M)$ following multinomial distribution to determine the number of neurons in each type, and then generate exponentially distributed Exp(1) for the non-zero element in each neuron. In the end, we normalize each type to one to make the solution on the manifold. It's also reasonable to use Dirichlet distribution or deterministic equal components as the data on the simplex.

For empirical investigation, we trained each global minimum with a 4-layer MLP with ReLU activations on MNIST dataset, with Kaiming-He initialization, SGD optimizer, batch size 100, and learning rate 0.05. Each minimum is trained with 10000 epochs. The widths for the network range from 15 to 100. For each reported value, we averaged the results for 10 independent realizations.

\subsection{Validation of Uniform Distribution}
\label{subsec:uniform}

\begin{figure}[htbp]
\centering
\begin{minipage}[c]{0.48\textwidth}
  \includegraphics[width=\textwidth]{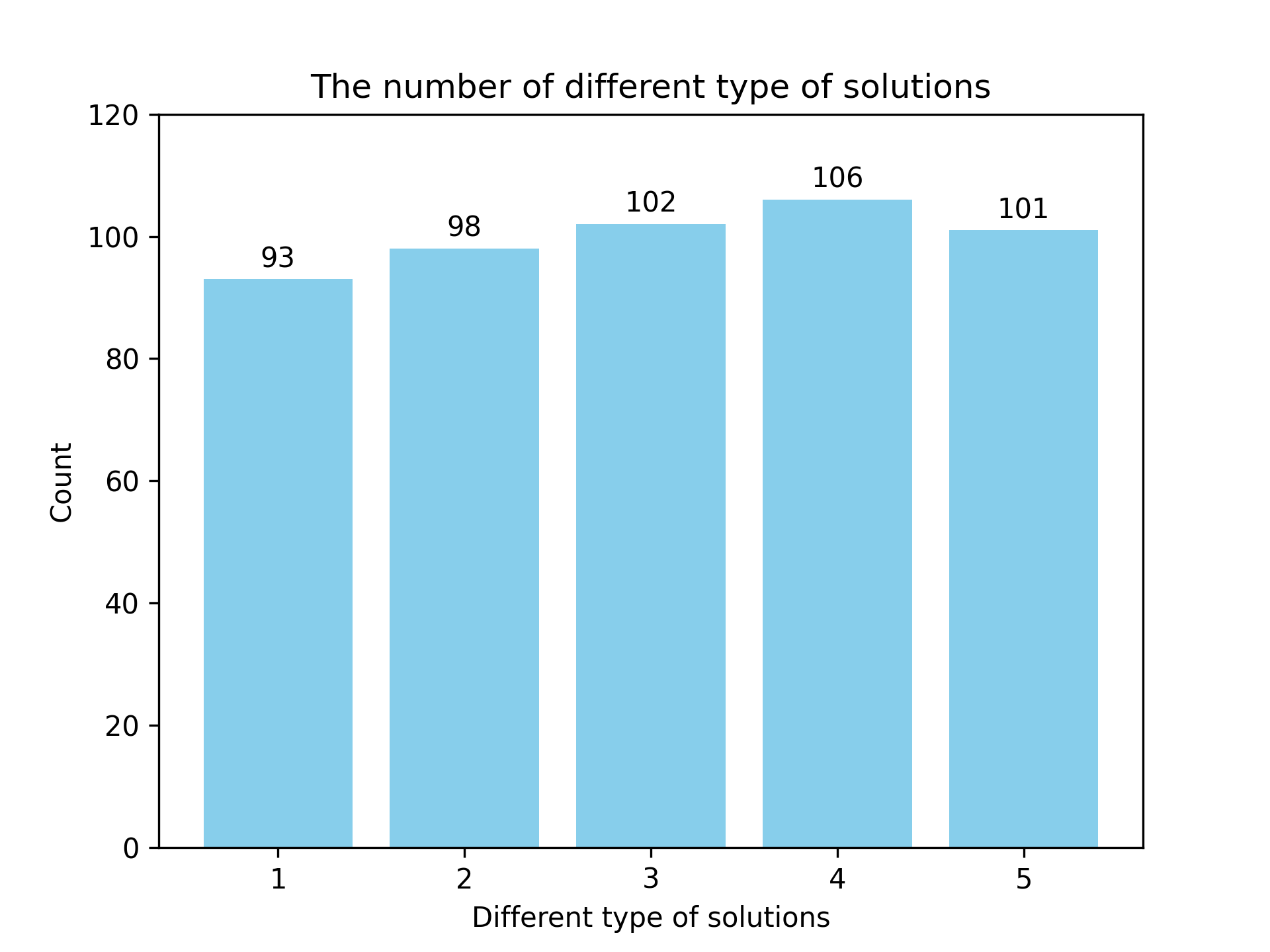}
  \subcaption{Type vector is multinomial}
  \label{subfig:multi}
\end{minipage}
\begin{minipage}[c]{0.48\textwidth}
  \includegraphics[width=\textwidth]{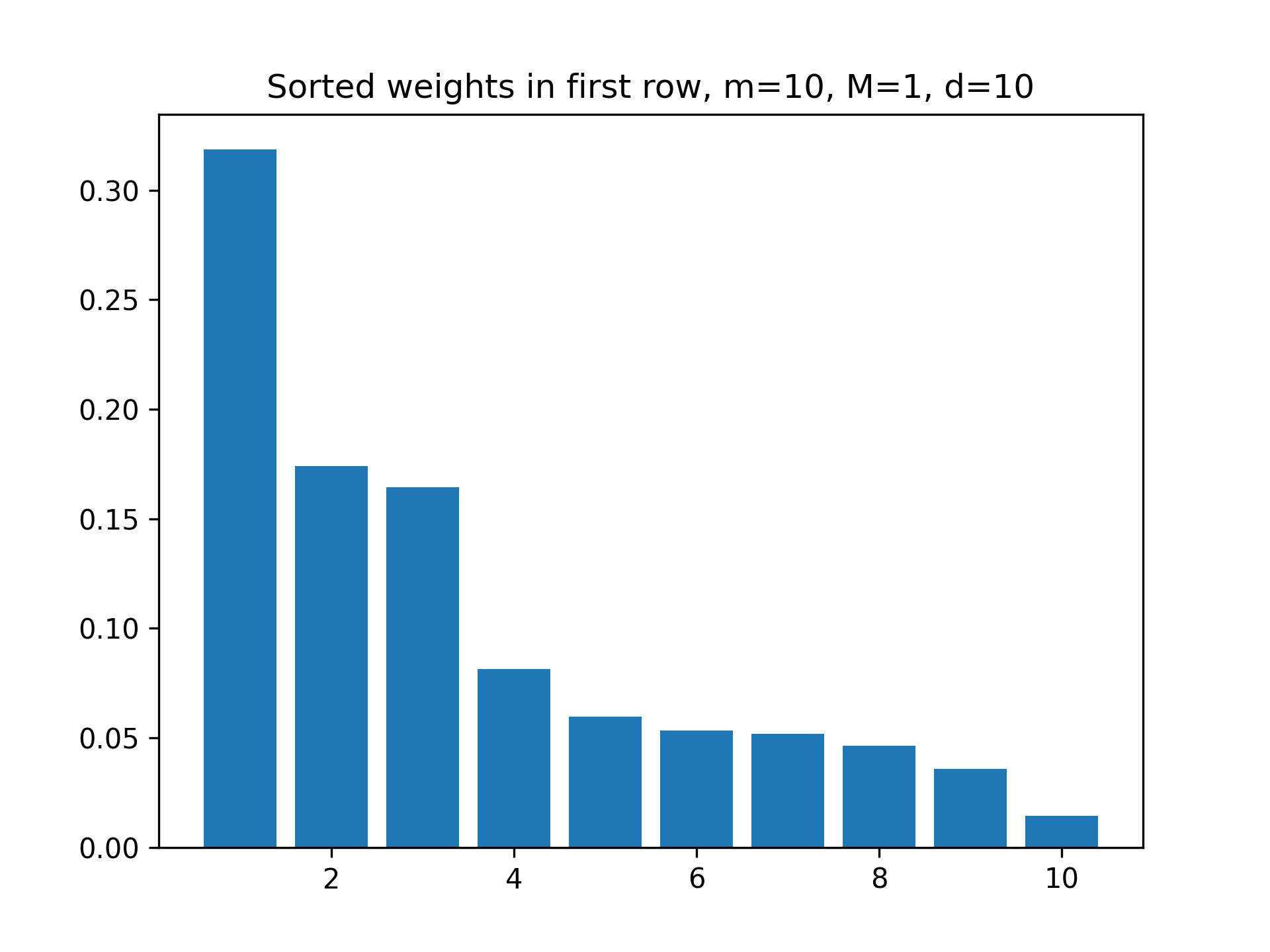}
  \subcaption{Each type is uniform}
  \label{subfig:simplex}
\end{minipage}
\caption{Evidence for Assumption \ref{assu:uniform}}
\end{figure}

In this subsection, we show some experiments validating our Assumption \ref{assu:uniform}. In Figure \ref{subfig:multi}, our setting is $m = 6,M=5,d=5$. Therefore, each type vector must be a unit vector $e_j, j \in [5]$. If $\alpha \sim \text{Multi}(1;1/5,1/5,1/5,1/5,1/5)$, then the probability of $\alpha = e_j$ is equally 1/5. We run 500 independent experiments and obtain 500 GD solutions, and we count the number of $\alpha = e_j$. As can be seen in the figure, the number of 5 different type vectors are almost the same, which confirms the validity of the multinomial distribution assumption and also corroborates that each neuron indeed gets assigned to different classes with equal probability.

In Figure \ref{subfig:simplex}, our setting is $m = 10,M = 1,d=10$. Therefore, there is only one teacher neuron and all neurons is type 1. We plotted the weights of the first column of W, sorted from large to small (as other columns are all zeros). It can be seen that the distribution of these elements is consistent with sampling from an exponential distribution, so according to Lemma \ref{lemma:exp} earlier, it is also sampling from a uniform distribution on the simplex. However, it's worth noting that based on our extensive experiments, this distribution pattern is not stable enough with the changes in learning rate, $m, M$, and $d$. Therefore, it is also reasonable to model the distribution on this simplex in other ways.

\subsection{Barrier Curve in Different Settings}
\label{subsec:morebarrier}
In this subsection we show some extra experiments in the simulation setting, giving comprehensive results for the behavior of the barrier of permuted minima and uniformly sampled solutions.

\begin{figure}[htbp]
\centering
\begin{minipage}[c]{0.49\textwidth}
  \includegraphics[width=\textwidth]{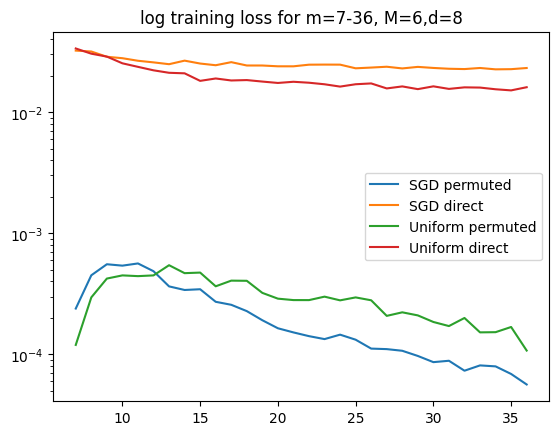}
  \subcaption{$M=6,d=8$}
\end{minipage}
\begin{minipage}[c]{0.49\textwidth}
  \includegraphics[width=\textwidth]{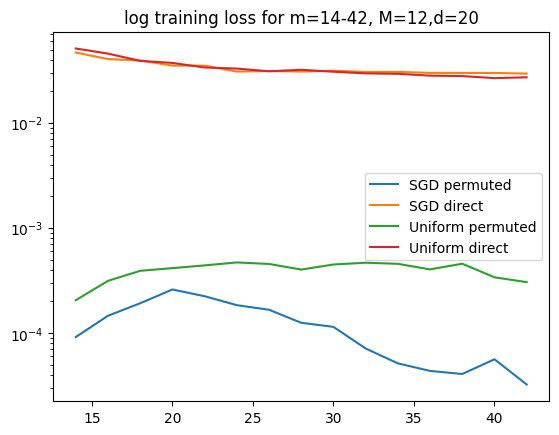}
  \subcaption{$M=12,d=20$}
\end{minipage}
\begin{minipage}[c]{0.5\textwidth}
  \includegraphics[width=\textwidth]{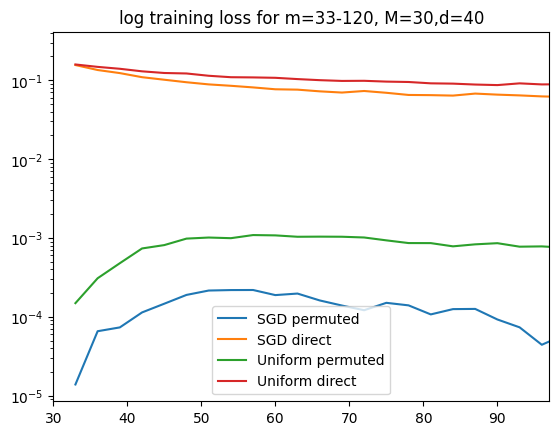}
  \subcaption{$M=30,d = 40$}
\end{minipage}
\caption{Barrier curve for different $m,M,d$. We increase $M$ and $d$ simultaneously. In each setting, we all can observe the trend of barrier first going up and then going down to 0. When $M$ is larger, barrier of permuted SGD solutions is way more lower than that of uniformly sampled solutions. This is partly due to the sparsity of solution. Each data point here is an average of 20 independent realizations.}
\label{fig:barrier-add}
\end{figure}

\begin{figure}[htbp]
\centering
\begin{minipage}[c]{0.43\textwidth}
  \includegraphics[width=\textwidth]{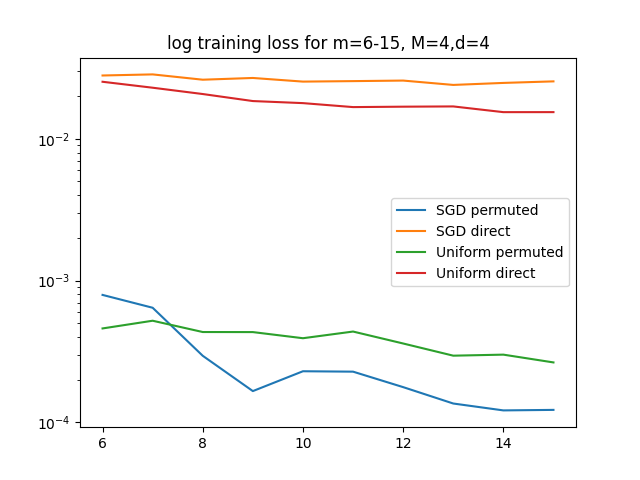}
  \subcaption{$d=4$}
\end{minipage}
\hfill
\begin{minipage}[c]{0.43\textwidth}
  \includegraphics[width=\textwidth]{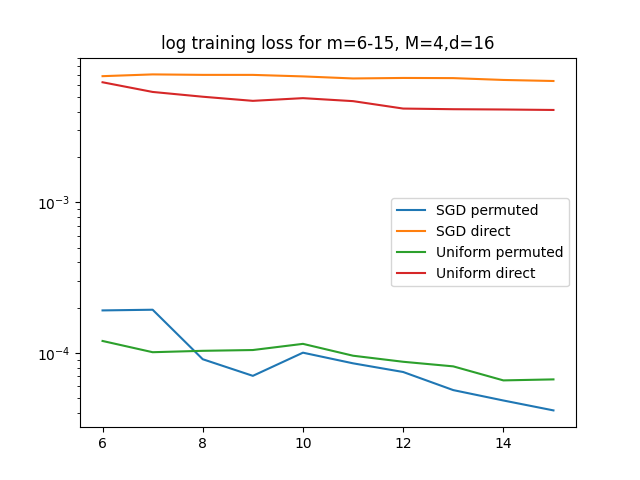}
  \subcaption{$d=16$}
\end{minipage}

\begin{minipage}[c]{0.43\textwidth}
  \includegraphics[width=\textwidth]{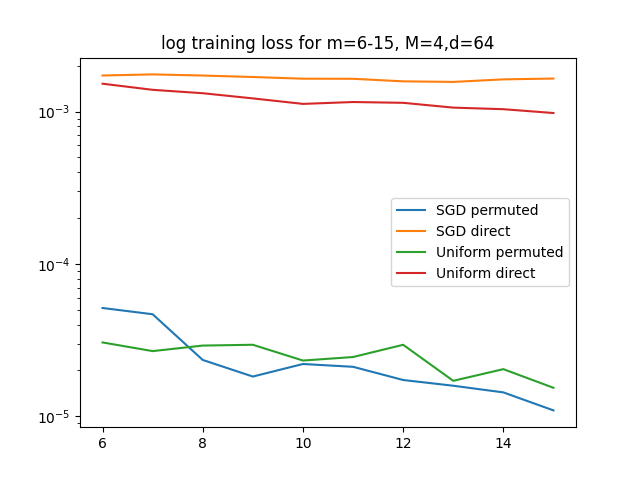}
  \subcaption{$d=64$}
\end{minipage}
\hfill
\begin{minipage}[c]{0.43\textwidth}
  \includegraphics[width=\textwidth]{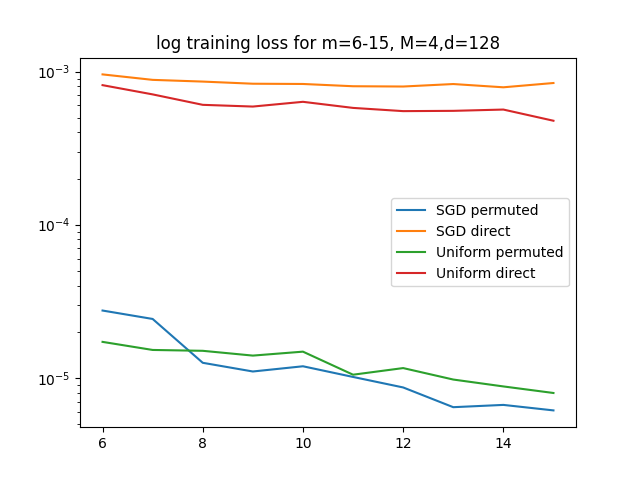}
  \subcaption{$d=128$}
\end{minipage}
\begin{minipage}[c]{0.5\textwidth}
  \includegraphics[width=\textwidth]{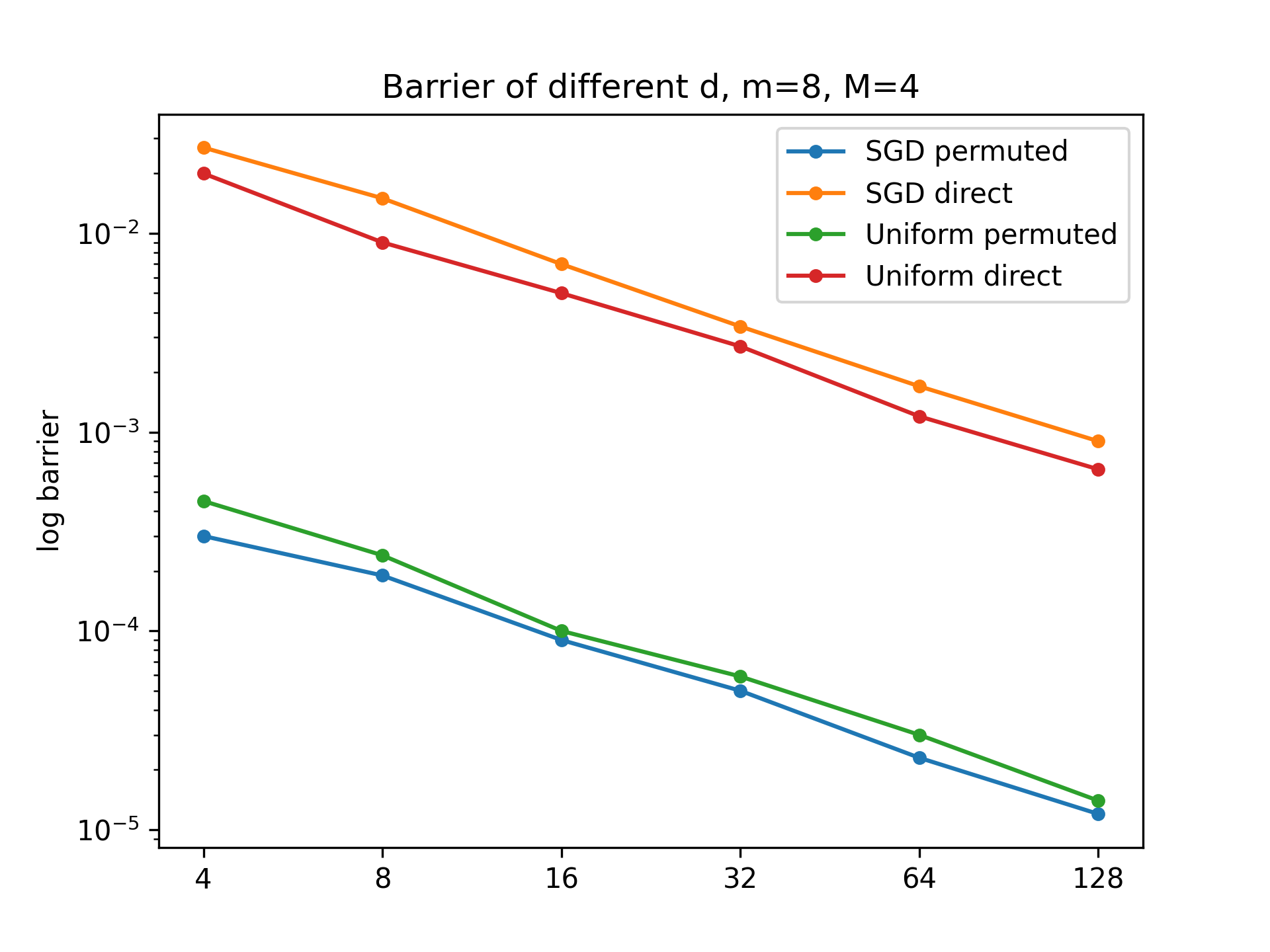}
  \subcaption{Profile for different $d$ at $m=8$}
  \label{fig:d-profile}
\end{minipage}

\caption{Barrier curve for different $d$ with fixed $m \in [6,15], M = 4$. (a) - (d) give the curve for different $m$ with increasing $d$, and (e) gives a profile at $m = 8,M=4$ with exponentially increasing $d$ from $2^2$ to $2^7$. We can observe that the barrier is decreasing when $d$ is going up, while the gap between the permuted barrier and the direct barrier remains almost the same.}
\label{fig:barrier-d}
\end{figure}

\begin{figure}[htbp]
\centering
\begin{minipage}[c]{0.32\textwidth}
  \includegraphics[width=\textwidth]{Additional_plots/Gap64.png}
  \subcaption{$M=4$}
\end{minipage}
\hfill
\begin{minipage}[c]{0.32\textwidth}
  \includegraphics[width=\textwidth]{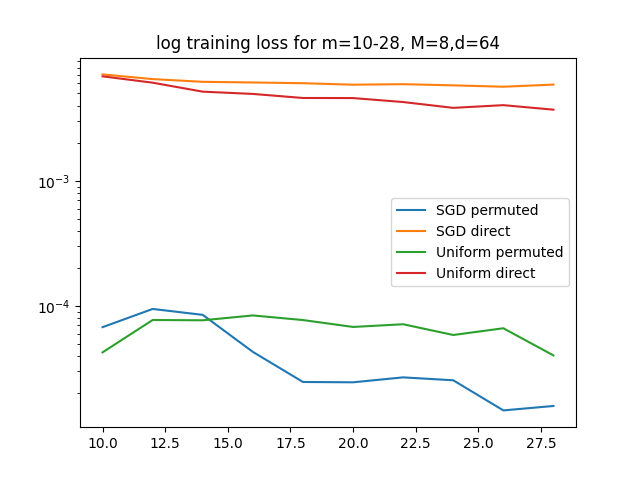}
  \subcaption{$M=8$}
\end{minipage}
\hfill
\begin{minipage}[c]{0.32\textwidth}
  \includegraphics[width=\textwidth]{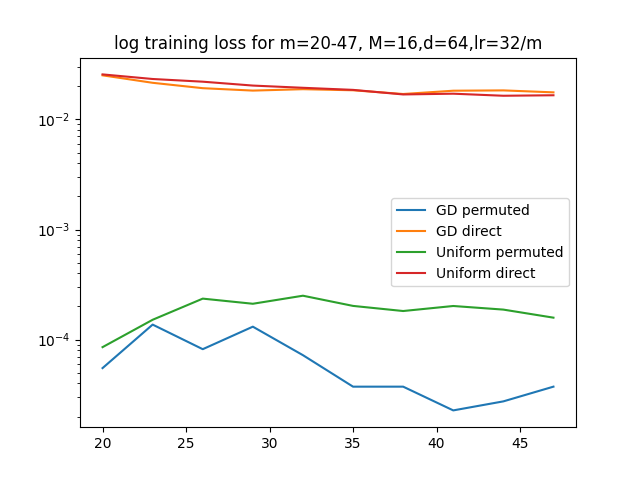}
  \subcaption{$M=16$}
\end{minipage}

\caption{Barrier curve for different $M$ with fixed $d = 64$. We can observe that the barrier is increasing when $d$ is going up, while the gap between the permuted barrier and direct barrier remains also becomes larger.}
\label{fig:barrier-M}
\end{figure}

\begin{figure}[htbp]
    \centering
    \includegraphics[width = 0.6\textwidth]{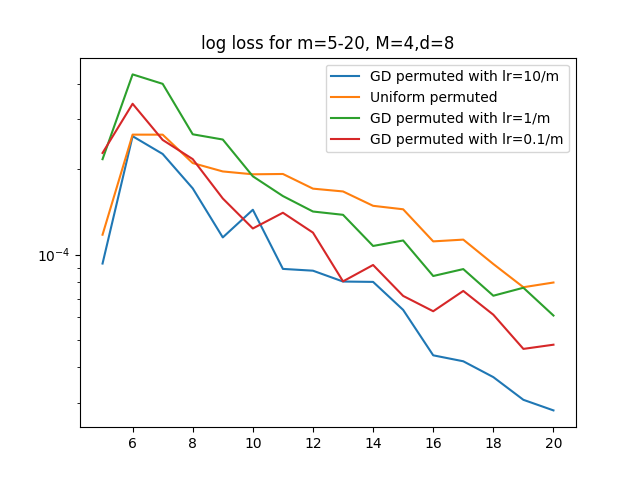}
    \caption{Barrier of two permuted minima obtained by GD of different learning rates. The barrier first goes up and then goes down with increasing learning rate. Each data point is an average of 50 simulations.}
    \label{fig:barrier-lr}
\end{figure}

\subsection{Peak Position}
In our theoretical setting, we obtained the result that the peak of barrier curve occurred at $m = 2M$. Following the setting of empirical investigation in \ref{subsec:detail}, we studied the elements affecting the peak position empirically. 

\begin{figure}[htbp]
    \centering
    \includegraphics[width = 0.55\textwidth]{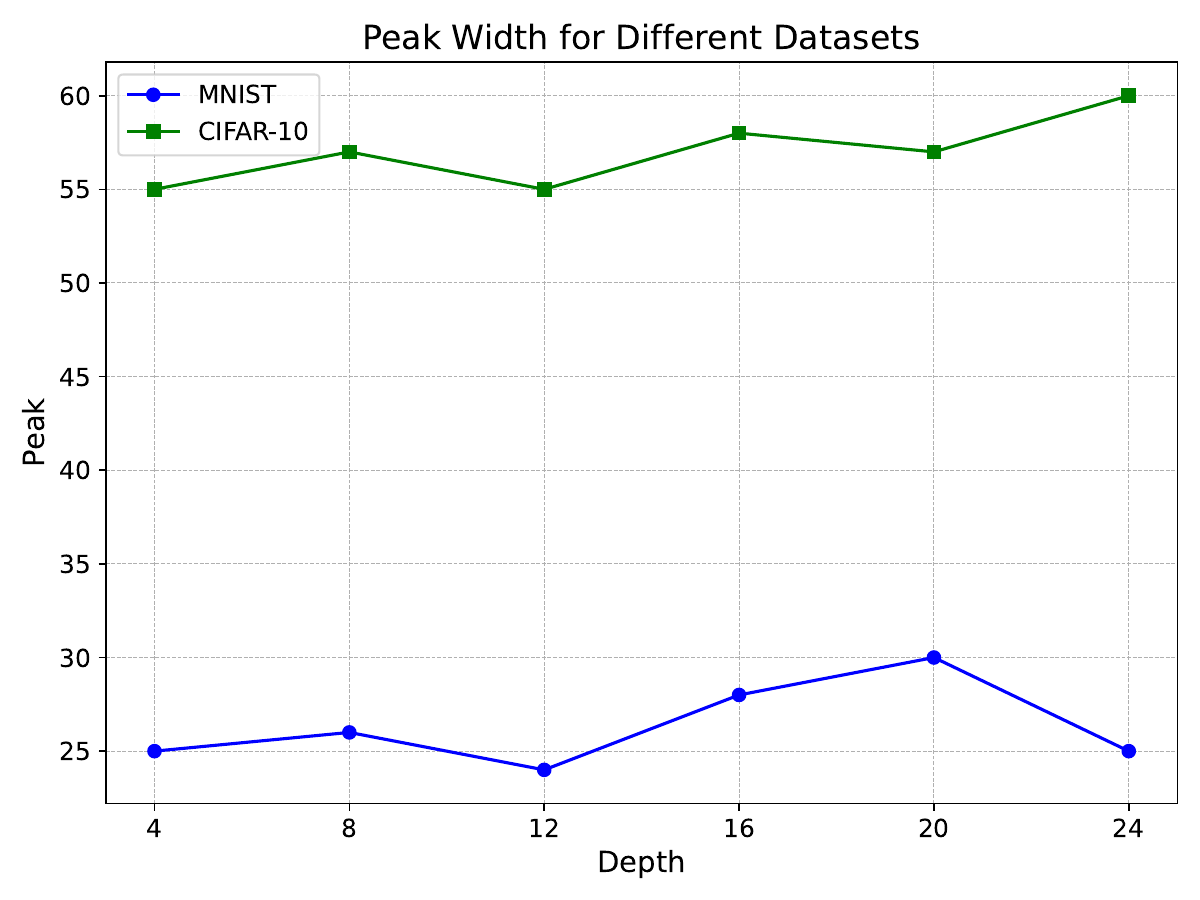}
    \caption{The peak widths for models trained on MNIST and Cifar10 with different depths.}
    \label{peak_depth}
\end{figure}

In Figure \ref{peak_depth}, we found that the peak position doesn't change according to depths, but it got bigger as our dataset became more complex. It aligns with our theoretical analysis in that the number of teacher neurons quantifies the difficulty of the task, which can be modeled by the complexity of the dataset.

\subsection{Cifar10 Experiments}
Figure \ref{interpolation} shows the barrier peak phenomenon of CNN trained on Cifar10. The result can also be verified by Figure 2 (left) provided in \cite{entezari2021role}.

\begin{figure}[htbp]
\minipage{0.33\textwidth}
  \includegraphics[width=\linewidth]{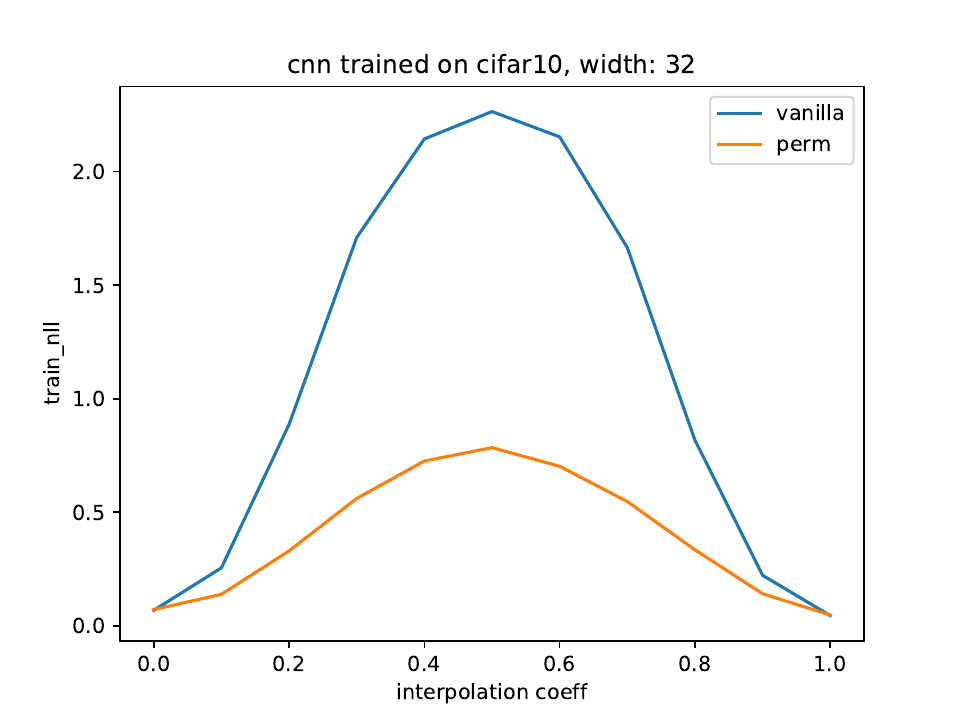}
\endminipage\hfill
\minipage{0.33\textwidth}
  \includegraphics[width=\linewidth]{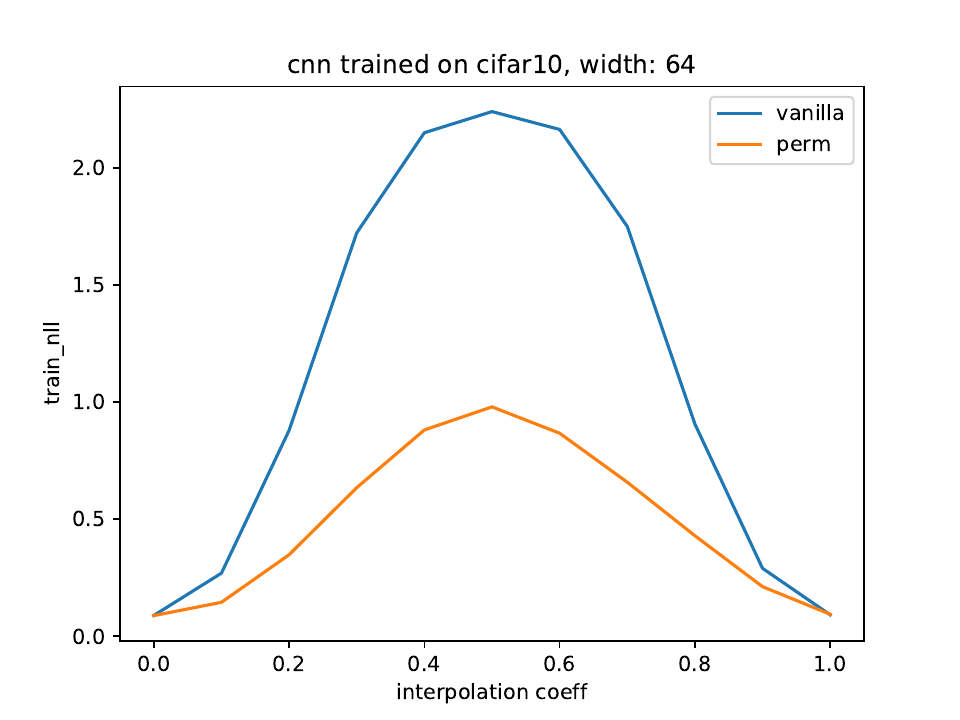}
\endminipage\hfill
\minipage{0.33\textwidth}%
  \includegraphics[width=\linewidth]{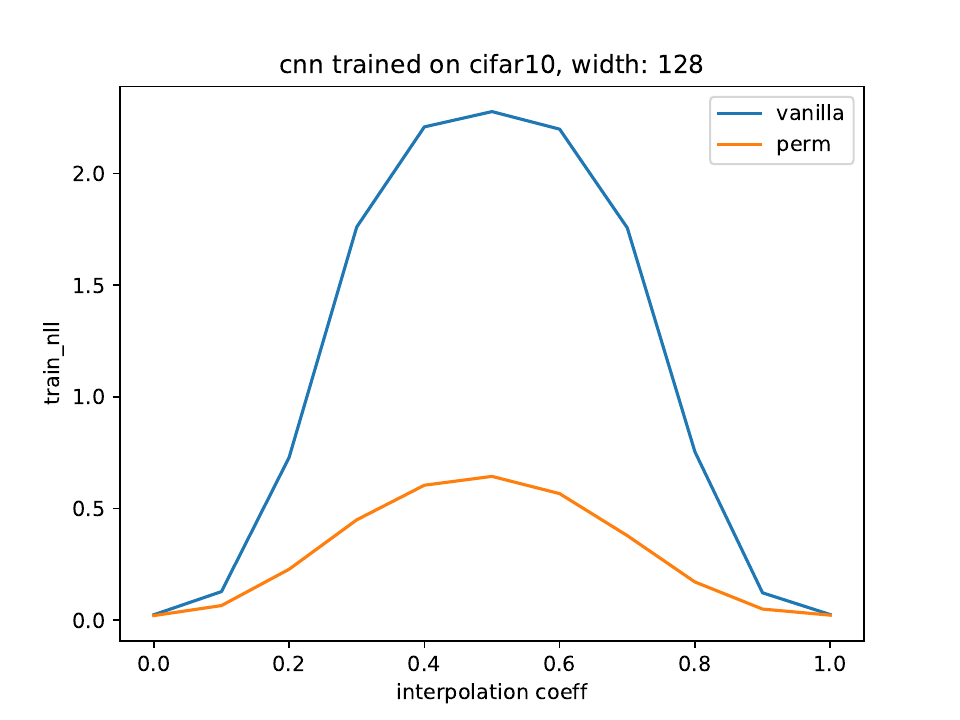}
\endminipage

\caption{Interpolation NLL plot under different widths on CIFAR10. The network is CNN, with depth 16, optimizer Adam and learning rate 0.005. The barrier goes up and then goes down as the width increases, indicating the existence of a peak.}
\label{interpolation2}
\end{figure}

\end{document}